\documentclass[square,numbers]{article}

\usepackage[preprint]{neurips_2024}

\usepackage[utf8]{inputenc} % allow utf-8 input
\usepackage[T1]{fontenc}    % use 8-bit T1 fonts
\usepackage{hyperref}       % hyperlinks
\usepackage{url}            % simple URL typesetting
\usepackage{booktabs}       % professional-quality tables
\usepackage{amsfonts}       % blackboard math symbols
\usepackage{nicefrac}       % compact symbols for 1/2, etc.
\usepackage{microtype}      % microtypography
\usepackage{xcolor}         % colors
\usepackage{times}
\usepackage{soul}
\usepackage{graphicx}
\usepackage{amsmath}
\usepackage{amsthm}
\usepackage{mdframed}
\usepackage{amssymb}
\usepackage{algorithm}
\usepackage{subfigure}
\usepackage{multirow}
\usepackage{wrapfig}
\usepackage{colortbl}
\usepackage{mathtools}
\usepackage{ulem}
\usepackage{algpseudocode}

\newtheorem{theorem}{Theorem}
\newtheorem{definition}{Definition}
\newtheorem{assumption}{Assumption}
\newtheorem{proposition}{Proposition}

\title{SSL Framework for Causal Inconsistency between Structures and Representations}

\author{%
  Hang Chen \\
  School of Computer Science and Technology\\
 Xi'an Jiaotong University\\
  \texttt{albert2123@stu.xjtu.edu.cn} \\
  \And
   Xinyu Yang\\
  School of Computer Science and Technology\\
 Xi'an Jiaotong University\\
  \texttt{yxyphd@mail.xjtu.edu.cn} \\
  \AND
    Keqing Du\\
  School of Computer Science and Technology\\
 Xi'an Jiaotong University\\
  \texttt{dukeqing@stu.xjtu.edu.cn} \\
  \And
  Wenya Wang\thanks{Corresponding author}\\
  College of Computing and Data Science\\
 Nanyang Technological University\\
  \texttt{wangwy@ntu.edu.sg} \\
}

\begin{document}

\maketitle

\begin{abstract}
The cross-pollination between causal discovery and deep learning has led to increasingly extensive interactions. It results in a large number of deep learning data types (such as images, text, etc.) extending into the field of causal discovery, and a multitude of deep learning tasks have begun to utilize causal discovery to explore the internal causal structure and causal representation of data. In this paper, we first identified that a complex data type, ``Indefinite Data", has conflicts between causal relationships expressed by the causal structure and causal representation generated by deep learning models, a phenomenon referred to as causal inconsistency. We thoroughly analyzed related work to explain why only Indefinite Data exhibits causal inconsistency while other data types do not. Furthermore, to alleviate causal inconsistency, we proposed a self-supervised learning (SSL) framework based on intervention, hoping to provide more causal information from different intervention views to promote consistency between structure and representation. Extensive experiments have shown that the SSL framework enhances causal consistency and can further improve causal structure and representation learning performance. Additionally, we extended the SSL framework to three different downstream tasks and LLM instructions. The quantitative results of these applications all reflect the performance improvement brought about by causal consistency.
\end{abstract}

\section{Introduction}
\label{introduction}

As an increasing amount of causal knowledge is extended to various types of deep learning tasks~\citep{jerzak2022image,ribeiro2023high,zhang2023towards,bagi2023generative,HAMMOND2023103919}, considering the internal causal relationships within data has become a popular aspect in numerous downstream tasks, such as time series forecasting~\citep{li2021causal,li2023transferable,chatziparaskevas2024generative}, dialogue generation~\citep{feng2023less,wan2023dialogue,hu2021causal}, and action segmentation tasks~\citep{du2023casr,liu2024knowledge,chen2022causal}.

From the perspective of causal discovery, some of these tasks require capturing \textbf{causal structures}, such as time series forecasting, which necessitates capturing the spatiotemporal causal relationships among multiple variables~\citep{runge2023causal,assaad2022survey}. Some tasks require generating \textbf{causal representations}, such as dialogue generation, which considers the representation of utterances which enables inferring underlying causal relationships in the context~\citep{wan2023dialogue,hu2021causal}. Some tasks require both structure and representation~\citep{du2023casr,chen2023affective}. We formalize such a causal discovery model as:
\begin{equation}
    \hat{X},\hat{\mathcal{G}}= CausalModel(X)
    \label{eqtcausalmodel}
\end{equation}
where $X$ represents the set of input causal variables, $\hat{\mathcal{G}}$ denotes the estimated causal structure (causal graph) of $X$, indicating the causal relationships among the causal variables in $X$, and $\hat{X}$ represents the causal representation of $X$, entailing the information of underlying and abstract causal relations from perceptible input, which can be subsequently used for classification, prediction, decision, etc. For instance, if the input is a dialogue containing $N$ utterances and the task is specified to conversation inference~\citep{poria2021recognizing}, then $\hat{\mathcal{G}}$ is equivalent to a directed acyclic graph (DAG) containing $N$ nodes, representing the causal relations between $N$ utterances, and $\hat{X} \in R^{N*d}$, where $d$ is the dimension of the representation, can represent the causal representation of each utterance.

Additionally, the expansion of causal inference in deep learning has led to the emergence of many new forms of causal data, one of which is referred to as \textbf{indefinite data}~\citep{chen2023review,chen2024causal}, which has two characteristics: a) Unlike single-structure data, the causal structure obtained from all samples is not necessarily equal, which we refer to as \textbf{multi-structure data}; b) Unlike some simple variables that inherently have a numerical form (such as age, blood pressure, temperature, etc.), the causal variables of this data are non-numerical, unstructured \textbf{complex variables} (such as text, video, audio, etc.).

We find that when indefinite data is used in the framework shown in Equation \ref{eqtcausalmodel}, the causal relationships implied by the generated $\hat{\mathcal{G}}$ and $\hat{X}$ are significantly conflicting while other data types as input do not, a phenomenon we refer to as causal inconsistency. 
To the best of our knowledge, the causal inconsistency in indefinite data has not been identified. In Section~\ref{secconsistency}, to dive into the causal inconsistency, we provide an in-depth analysis of why causal inconsistency only arises in the indefinite data. 

To resolve the inconsistency, in Section~\ref{secmethod}, we propose a self-supervised learning (SSL) framework, taking intervention measures into account to optimize the causal structures and representations towards a higher consistency. Specifically, different interventions can be regarded as different “views”, and the measures to gauge causal relationships are treated as “augmentations”. Extensive experiments in Section~\ref{secexperiments} demonstrate that the SSL framework can significantly mitigate causal inconsistency, thereby improving the estimation performance of the structure and representation of indefinite data. 

Additionally, in Section~\ref{secapplication}, we extend the SSL framework to downstream tasks and large language models (LLMs). Quantitative experimental results demonstrate that, after improving consistency, latent causal relationships in complex data like indefinite data are also unearthed, reflected in the significant enhancement provided for these downstream applications that require causal knowledge.

In summary, for causal learning on indefinite data, we have provided a series of novel and thorough contributions: we discover causal inconsistency, analyze the reasons for this inconsistency, propose the SSL framework to enhance consistency, and conduct extensive experiments over several application scenarios. These contributions lead us to believe this work can provide potential insights and broader impacts for causal learning on complex data.

\section{Preliminaries}
\label{secpreliminaries}
\subsection{Input-Output Framework}
\label{secinputoutputframework}
\begin{figure}
    \centering
    \includegraphics[width=0.7\linewidth]{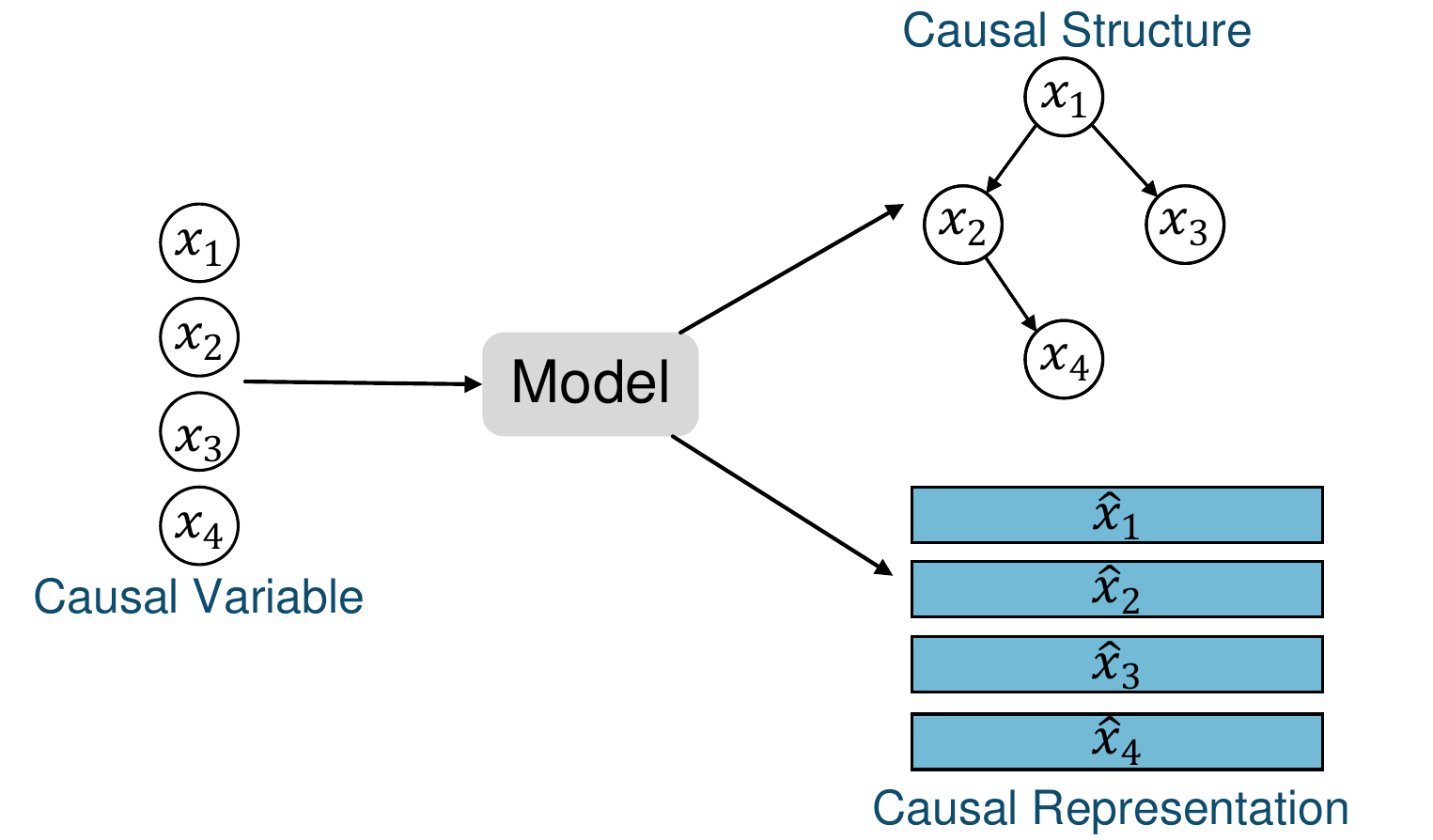}
    \caption{Input-output Framework of interest.}
    \label{figinputoutputframewrok}
 \end{figure}
 
According to Equation~\ref{eqtcausalmodel}, the input-output framework we are focusing on is shown in Figure~\ref{figinputoutputframewrok}. Suppose $X_{i}$ is any sample in the dataset, which contains $N$ causal variables, i.e., $X_{i}=\{x_{i, 1}, ..., x_{i, N}\}$.  Through a specific causal model, it generates two outputs: the causal structure $\hat{\mathcal{G}}_{i}$ and the causal representation $\hat{X}_{i}$. Here, $\hat{\mathcal{G}}_{i}$ represents the causal relationships between variables, existing in the form of a DAG, also known as a causal graph. $\hat{X}_{i}\in R^{N*d}$ represents $N$ $d$-dimensional representations, which are endowed with causal mechanisms through the causal model (for example, the structural causal model (SCM) is a popular causal mechanism), and are used for subsequent classification, generation, decision-making, etc.

\subsection{Categories in Structures and Representations}
Following the settings in~\citep{chen2023review}, we have classified the data. Structurally, there are single-structure data and multi-structure data~\citep{lowe2022amortized}. In terms of representation, there are simple variables and complex variables~\citep{chen2024causal}. We provide specific descriptions of these four types as follows:

\textbf{Single-structure data}: All samples in a dataset share a single causal structure. That is, for any sample $X_{i}$, the correct model should estimate a completely identical $\hat{\mathcal{G}}$.

\textbf{Multi-structure data}: There are $M$ causal structures ($M>1$) in the entire dataset. That is, for any sample $X_{i}$, the correct model should estimate one of the $M$ structures $\hat{\mathcal{G}}_{m}\in \{\hat{\mathcal{G}}_{1},...\hat{\mathcal{G}}_{M}\}$.

\textbf{Simple variables}: Variables that exist in samples in a fixed numerical form, such as blood pressure, temperature, age, etc. Most of the time, the values of such variables themselves are used as causal representations without the need for additional calculations, i.e., in the input-output framework, $\hat{X}=X$.

\textbf{Complex variables}: Variables that do not have a fixed numerical form in samples, such as text, video, and data in other modalities. The input $X$ is some initial representation of it (such as embeddings), and the correct model should apply a causal mechanism to it, therefore usually $\hat{X}\neq X$\footnote{In the scenario of simple variables, with small probability, it is necessary to calculate a representation for such variables that is not equal to the original value. In this case, they are treated as complex variables.}. 

\begin{figure*}
  \centering
  \subfigure[Single-structure data]{
    \includegraphics[width=0.45\textwidth]{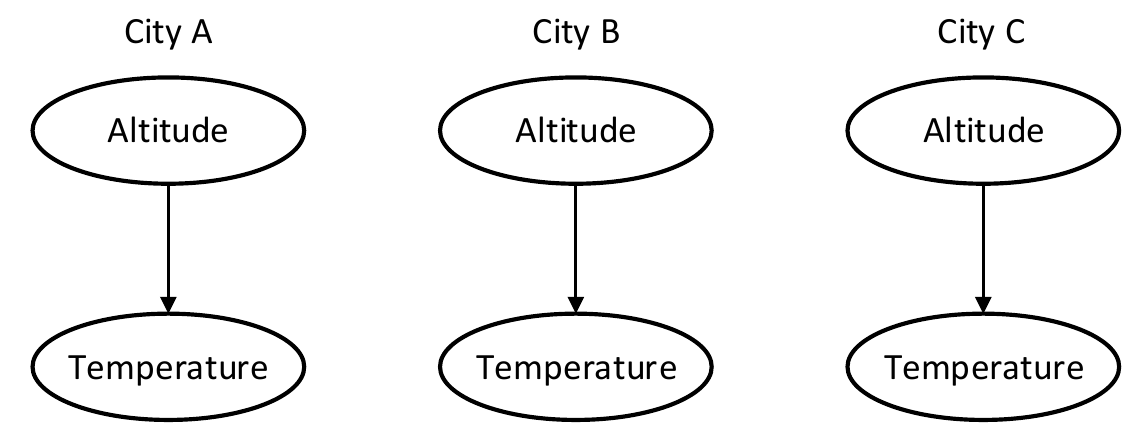}}
  \subfigure[Multi-strucutre data]{
    \includegraphics[width=0.45\textwidth]{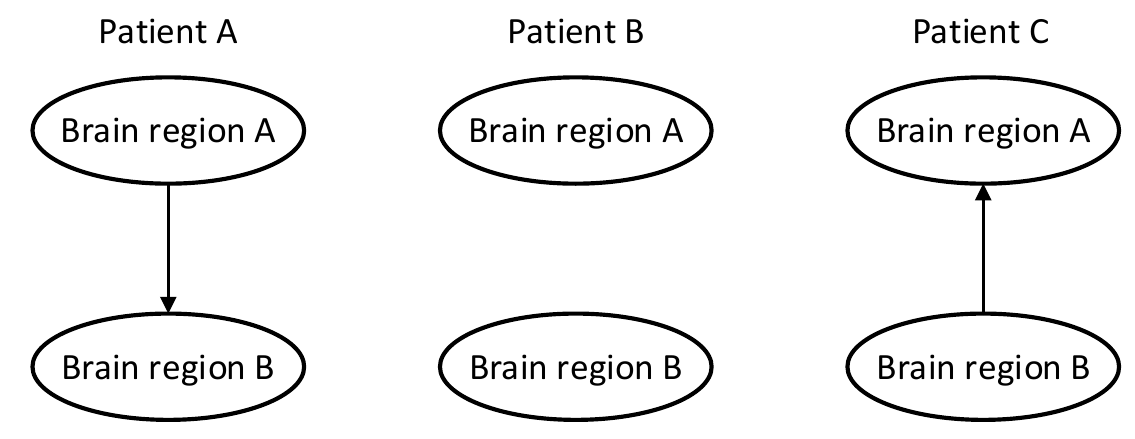}}
  
  \caption{Comparison between single-structure data and multi-structure data.}
  \label{figcomparisonofstructure}
\end{figure*}

It's worth noting that multi-structure data does not imply that each sample corresponds to multiple causal structures. In Figure~\ref{figcomparisonofstructure}, we list three samples of single-structure data and multi-structure data respectively. In single-structure data, we observe the influence of the variable ``Altitude'' on the variable ``Temperature''. As a physical law, it does not change with the sample. However, in multi-structure data, we observed the influence of the variable ``Brain region A'' on the variable ``Brain region B''. Since different patient samples have different symptoms, these symptoms cause different electrical signal response rules between brain areas, i.e., different causal structures. 

\begin{table}
  \centering
  \resizebox{0.7\linewidth}{!}{
  \begin{tabular}{c|c c c c c}
    \hline
    \textbf{Category}&\textbf{Variables}&$X$&\textbf{$\hat{X}$}&\textbf{Dimension}(d)\\
    \hline
    \multirow{2}{*}{simple variable}&Age&25 & 25 &$d=1$\\
    &Voltage &2&2 &$d=1$\\
    \hline
    \multirow{2}{*}{complex variable}&Token&tensor&tensor&$d>1$\\
    &Frame &tensor&tensor&$d>1$\\
    \hline
\end{tabular}}
\caption{Comparison between simple variables and complex variables}
  \label{tabvariables}
\end{table}
Additionally, we have also compared simple variables and complex variables in Table~\ref{tabvariables}. For simple variables, we select a sample containing age and a sample containing voltage. For complex variables, we select a sample containing text and a sample containing video. Obviously, for simple variables, $\hat{X}=X$, so effective verification can be carried out for causal representation (for example, a statistical approach like independent components analysis (ICA)). However, for complex variables, $\hat{X}\neq X$, and due to the high-dimensional continuous representation, it is difficult to verify causal representation through these statistics-based methods.

Therefore, in this paper, we default to using neural networks transferring representations to relationships to validate the causal representation of complex variables. Specifically, for any two variables in any sample $X_{s}$ arranged in order $<x_{s,i}, x_{s,j}>$, if there exists a causal relationship $x_{s,i}\rightarrow x_{s,j}$, we have label $Y_{<x_{s,i}, x_{s,j}>}=1$, otherwise, we set $Y_{<x_{s,i}, x_{s,j}>}=0$. Let $f_c$ be a causal classifier, e.g., an MLP followed by a sigmoid function, with the input being the causal representation of any two variables $<\hat{x}_{s,i}, \hat{x}_{s,j}>$, and it is supervised and trained according to the label $Y_{<x_{s,i}, x_{s,j}>}$. During validation, if $f_c(<\hat{x}_{s,i}, \hat{x}_{s,j}>) \in (0, 0.5]$, it indicates there is no relationship pointing from $x_{s,i}$ to $x_{s,j}$, and if $f_c(<\hat{x}_{s,i}, \hat{x}_{s,j}>) \in (0.5, 1)$, there is a relationship $x_{s,i}\rightarrow x_{s,j}$.

From above basic types of causal data, the new data paradigm named  
``indefinite data''~\citep{chen2023review} is defined as: 

\begin{definition}[Indefinite Data]
    The causal relationships exist in a dataset 
    $\mathbf{D} = \{X_{s}\}^{S}_{s=1}$ which has $S$ samples 
    and $M$ ($M > 1$) causal structures 
    ($\mathcal{G}=\{\mathcal{G}_{m}\}^{M}_{m=1}$). 
    Each structure is formalized as a graph, i.e., $\mathcal{G}_{m}=(X_{m}, \mathcal{E}_{m})$ (where $X_{m}$ stands for the nodes and $\mathcal{E}_{m}$ represents edges). 
    Hence, each sample $X_{s,m} \in \mathbb{R}^{N_{m} \times d}$  ($d > 1$)
    belongs to a causal structure  
    $\mathcal{G}_{m}$ and 
    consists of $N_{m}$ variables: 
    $X_{s, m}=\{x_{s,m,n}\}^{N_{m}}_{n_{m}=1}$. 
    $\hat{X}_{s,m} \in \mathbb{R}^{N_{m} \times d}$ 
    represents the causal representation of $X_{s,m}$. 
    \label{defcausalmodel}
  \end{definition}

Generally, the identifiability of causal relationships is guaranteed by acyclic constraints (e.g., NOTEARS~\citep{zheng2018dags}). However, for simplification, we assume that all indefinite data comply with the time order (indeed, in this paper, all the indefinite datasets adhere to this assumption).

\begin{assumption}[Causal Identifiability]
    The index of causal variables in indefinite data sample $X_{s,m}$ satisfies time order, defined as a linear order $\prec_{X_{s,m}}$. Let $\mathcal{J}_{X_{s,m}}$ be the index set of $X_{s,m}$. $\forall i,j \in \mathcal{J}_{X_{s, m}}$, if $i<j$, $x_{i} \prec_{X_{s,m}} x_{j}$.
    \label{hypindentifiability}
\end{assumption}
Therefore, the causal order can be regarded as a partial order, defined as $\preccurlyeq_{X_{s,m}}$ w.r.t. the time order $\prec_{X_{s,m}}$. That is, $\forall  x_{i} \prec_{X_{s,m}} x_{j}$, there must be $x_{i} \preccurlyeq_{X_{s,m}} x_{j}$. 

\subsection{Related Work}
\subsubsection{Causal Structure Learning}
\begin{figure}
    \centering
    \includegraphics[width=0.6\linewidth]{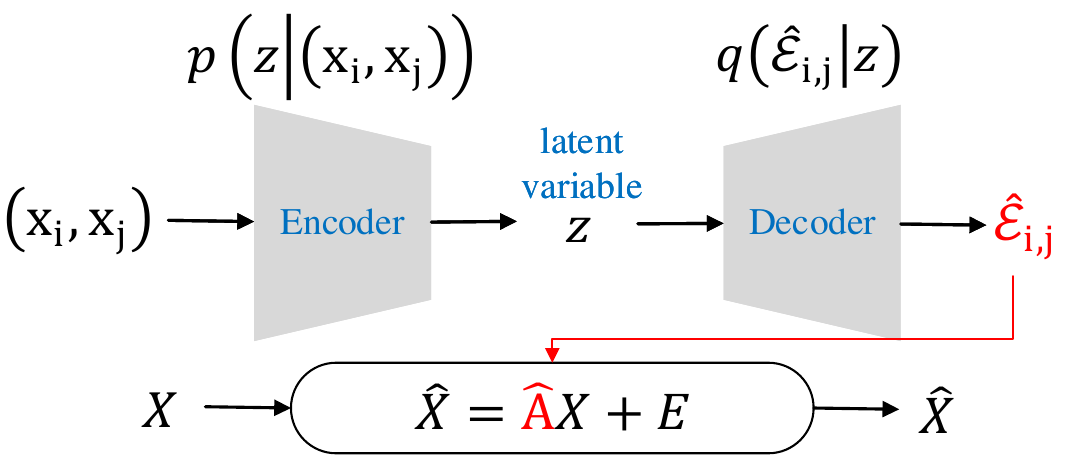}
    \caption{Generative framework to solve multi-structure and simple-variable data.}
    \label{figmultistructurecausalmodel}
 \end{figure}
In causal structure learning, causal variables are typically simple variables, i.e., in our input-output framework, the output $\hat{X}$ should equal $X$. 
Therefore, for single-structure data, a large body of work has demonstrated that accurate statistics between variables can be used to recover the causal structure~\citep{kalisch2007estimating,chickering2002optimal,hauser2012characterization,hoyer2008nonlinear}. For multi-structure data, it was initially treated as multiple single-structure problems~\citep{tank2021neural,peters2017elements}. However, this leads to a situation where a new model needs to be refitted whenever a new causal structure appears. 

Recently, to address the problem of multi-structure scenarios, existing methods typically employ amortized causal discovery approaches~\citep{lorch2022amortizedinferencecausalstructure,lowe2022amortized,huang2020causala,
dhir2020integrating,huang2020causalb,huang2019specific}. Specifically, multiple causal structures are viewed as a distribution, with each structure being a sample within it. An estimated adjacency matrix of the structure, $\hat{A}$, can be obtained through the generative model shown in Figure~\ref{figmultistructurecausalmodel}, and then the causal representation $\hat{X}$ is derived via the SCM. As simple variables, $\hat{X}$ is equivalent to $X$, allowing the reconstruction loss to be formulated as distance measure between $\hat{X}$ and $X$, i.e., $E_{q(\hat{\mathcal{E}}_{i,j}|z)}[log p(z|(x_i, x_j))]=distance(\hat{X}, X)$. 

\subsubsection{Causal Representation Learning}
\label{seccausalrepresentationlearning}
\begin{figure}
    \centering
    \includegraphics[width=0.6\linewidth]{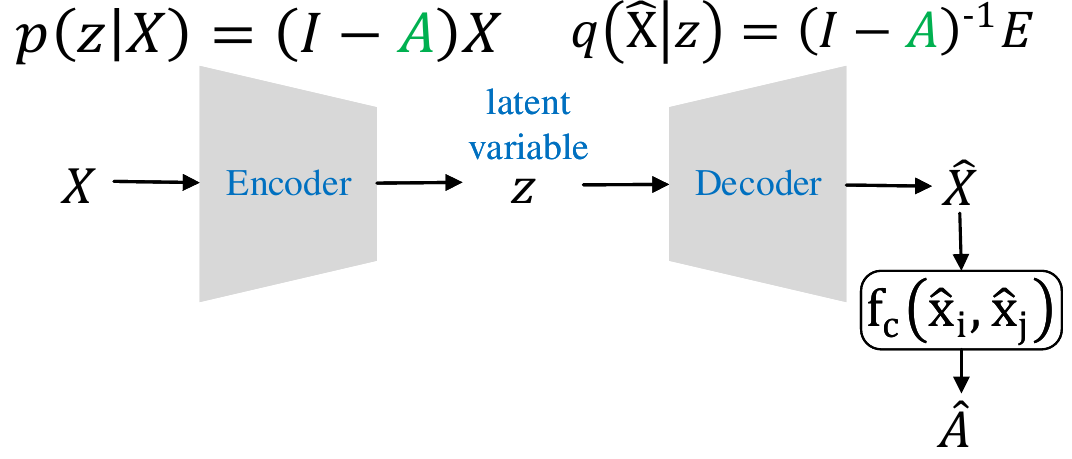}
    \caption{Generative framework to solve single-structure and complex-variable data.}
    \label{figcomplexvariablecausalmodel}
 \end{figure}
In causal representation learning, we only consider those cases involving complex variables, i.e., situations where $\hat{X}\neq X$. In these cases, it is usually accompanied by a single, fixed, and known\footnote{``Known" means that the causal structure $\mathcal{G}$ can be incorporated into the computation as a known quantity, both during the training and validation processes.} causal structure~\citep{fan2022debiasing,wu2022discovering,lv2022causality,jiang2022invariant}, i.e., in input-output framework, causal structure does not need to be outputted. 
 Under such an assumption, as shown in Figure~\ref{figcomplexvariablecausalmodel}, the noise term $E$ of the SCM can be treated as a latent variable, thus the encoder and decoder can be viewed as a back-and-forth mapping between $X$ ($\hat{X}$) and $E$ through known adjacency matrix $A$ ($\mathcal{G}$). Finally, the corresponding $\hat{A}$ is obtained from $\hat{X}$ through causal classifier $f_{c}$. Therefore, the reconstruction loss can be measured by the distance between $\hat{A}$ and the known structure $A$, that is, $E_{q(\hat{X}|z)}(log p(z|X))=distance(\hat{A}, A)$.

\subsubsection{Causal Discovery for Indefinite Data}
\label{secindefinitedatacausaldiscovery}
\begin{figure}
    \centering
    \includegraphics[width=0.8\linewidth]{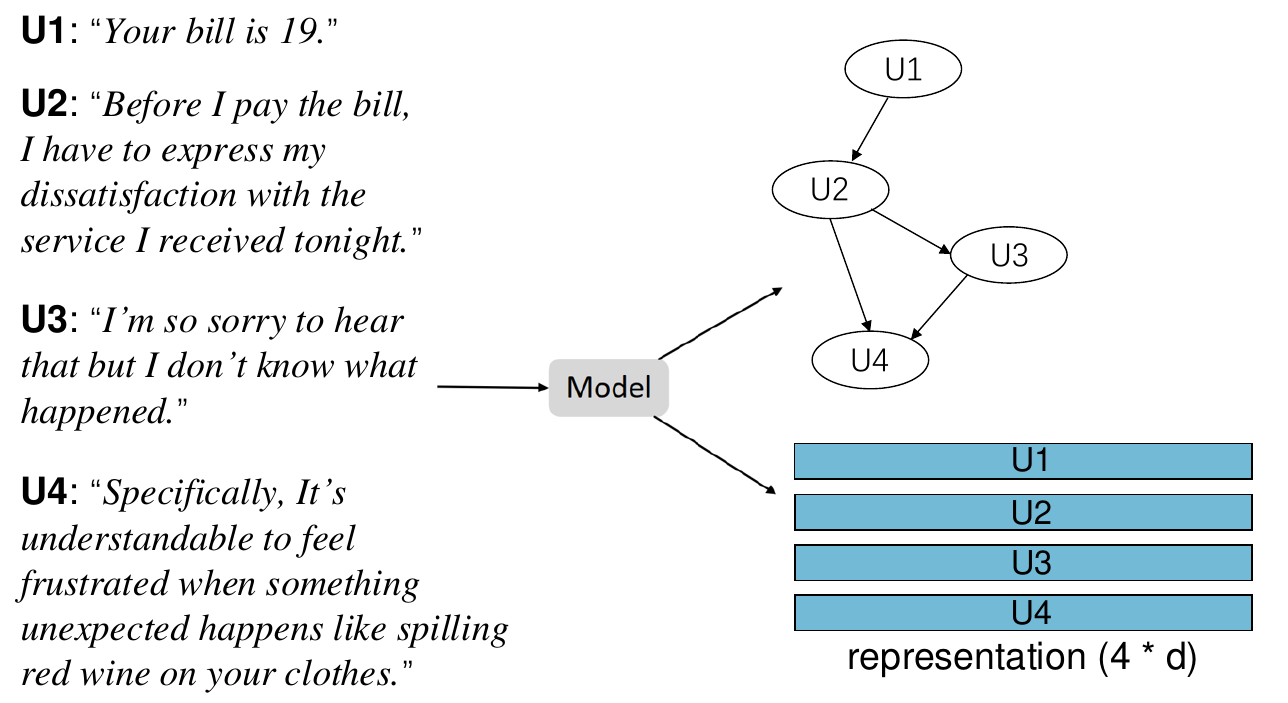}
    \caption{Sample from Causalogue dataset, applied in our input-output framework.}
    \label{figindefinitecase}
 \end{figure}
Causal discovery in indefinite data (CDID) is defined as the task that requires simultaneous learning of causal structure from multi-structure data and causal representation from complex variables~\citep{chen2024causal}. That is, in the input-output framework, for any input sample, the corresponding causal structure and causal representation need to be produced, and $\hat{X} \neq X$. In the indefinite dataset, causal structures are provided as labels (but cannot be treated as known quantities directly as in Section~\ref{seccausalrepresentationlearning}).
 \begin{figure}
    \centering
    \includegraphics[width=0.7\linewidth]{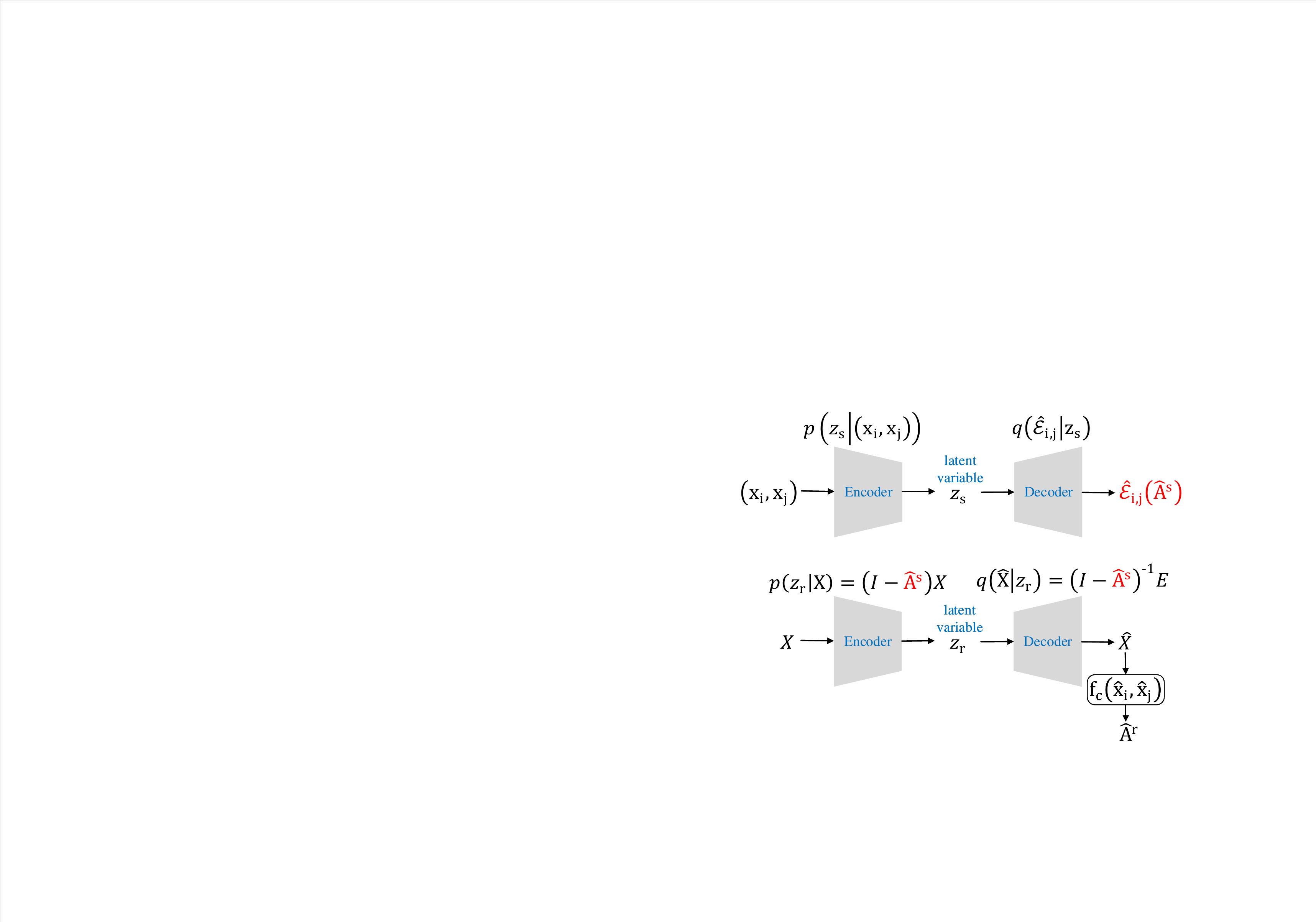}
    \caption{Generative framework to solve multi-structure and complex-variable data.}
    \label{figindefinitecausalmodel}
 \end{figure}
In Figure~\ref{figindefinitecase}, we use a sample from the Causalogue dataset~\cite{chen2024causal} to illustrate indefinite data that complies with Assumption~\ref{hypindentifiability}. This sample is a dialogue containing $4$ utterances, considered as $4$ causal variables. The task requires estimating the causal relationships of these $4$ utterances and their corresponding causal representation. The causal representation is validated through $f_c$. 

To simultaneously address the issues of multi-structure and complex variables in indefinite data, existing research has adopted a combination of the above two generative models as a new baseline model~\citep{chen2023learning,chen2023affective,chen2024causal}. As shown in Figure~\ref{figindefinitecausalmodel}, the estimated structure $\hat{A}^{s}$ is first obtained through the multi-structure generative model, and then input into the representation generative model to obtain the causal representation. Another estimated structure $\hat{A}^{r}$ is obtained from $\hat{X}$ through a causal classifier $f_c$. There are two reconstruction losses, calculated by measuring the distance between the two estimated structures and the ground truth $A$, i.e., $E_{q(\hat{\mathcal{E}}_{i,j}|z_s)}[log p(z_s|(x_i, x_j))]=distance(\hat{A}^{s}, A)$, $E_{q(\hat{X}|z_r)}(log p(z_r|X))=distance(\hat{A}^{s}, A)$. 

\section{Why Does Inconsistency Arise?}
\label{secconsistency}
 We define inconsistency as a conflict between the causal relationships implied by the causal representation $\hat{X}$ and estimated structure $\hat{\mathcal{G}}$. For instance, in $\hat{\mathcal{G}}$, we obtain $x_{1} \rightarrow x_{2}$, but $f_{c}(x_{1},x_{2})$ lags far from $1$. Formally, let the adjacency matrix of the causal structure $\hat{\mathcal{G}}$ be $\hat{A}^{s}$, and we estimate $\hat{A}^{r}_{i,j}=f_{c}(x_{i},x_{j})$ through the causal representation. Hence, the causal inconsistency can be quantified by $MSE(\hat{A}^{s},\hat{A}^{r})$. 

\begin{table}
  \centering
\resizebox{1\linewidth}{!}{
  \resizebox{0.45\linewidth}{!}{
 
		\centering
            \subtable[Datasets]{
		\begin{tabular}{c|cc}
        \hline
		dataset&single-structure&multi-structure\\
            \hline
            simple variable& Arrhythmia&Netsim\\
            complex variable&CMNIST-75sp&Causalogue\\
            \hline
		\end{tabular}}}
 \resizebox{0.45\linewidth}{!}{
 
		\centering
            \subtable[Approach]{
		\begin{tabular}{c|cc}
        \hline
		approach&single-structure&multi-structure\\
            \hline
            simple variable& PC,SGS&ACD, AVICI\\
            complex variable&DAG-GNN, CVAE&CAE, biCD\\
            \hline
		\end{tabular}}}
 }
 \caption{Datasets and approaches in 4 combinations of data types.}
  \label{tabdifferentdatatype}
\end{table}
\begin{table*}
  \centering
  \resizebox{0.9\textwidth}{!}{
  \begin{tabular}{c|ccc|ccc|ccc|ccc}
    \hline
    \bf Methods&\multicolumn{3}{c}{\textbf{Arrhythmia}} \vline&\multicolumn{3}{c}{\textbf{Netsim}}\vline&\multicolumn{3}{c}{\textbf{CMNIST-75sp}} \vline&\multicolumn{3}{c}{\textbf{Causalogue}} \\
    &\textbf{stru}&\textbf{rep}&\textbf{inco}&\textbf{stru}&\textbf{rep}&\textbf{inco}&\textbf{stru}&\textbf{rep}&\textbf{inco}&\textbf{stru}&\textbf{rep}&\textbf{inco}\\
    \hline
    PC&0.08&-&0.05&0.13&-&0.09&-&-&-&-&-&-\\
    SGS&0.02&-&0.04&0.09&-&0.09&-&-&-&-&-&-\\
    ACD&-&-&-&0.08&0.03&0.07&-&-&-&0.22&0.29&0.38\\
    AVICI&-&-&-&0.08&0.05&0.08&-&-&-&0.31&0.29&0.35\\
    DAG-GNN&0.02&0.03&0.06&0.17&0.05&0.08&0.05&0.15&0.07&0.33&0.35&0.37\\
    CVAE&-&0.06&0.06&-&0.02&0.10&-&0.18&0.08&-&-&-\\
    CAE&-&-&-&-&-&-&0.08&0.16&0.09&0.29&0.41&0.42\\
    biCD&0.08&0.09&0.08&0.11&0.06&0.10&0.06&0.14&0.09&0.34&0.38&0.29\\
    \hline
  \end{tabular}}
  \caption{Cross experiments of causal discovery models on different types of datasets.  For each dataset, we investigate three metrics: ``stru", ``rep", and ``inco". ``Stru" represents the difference between the estimated causal structure and the ground-truth causal structure. Specifically, stru=$MSE(\hat{A}^{s},A)$. ``Rep" represents the difference between causal representation and ground-truth representation. We obtain the adjacency matrix $A$ through the ground-truth causal structure, and then rep=$MSE(\hat{A}^{r},A)$. ``Inco" represents the causal inconsistency, i.e., inco=$MSE(\hat{A}^{s},\hat{A}^{r})$. All indicators are within the range $[0,1]$, and the closer to $1$, the worse the performance.}
  \label{tabinconsistency}
\end{table*}

In our experiments, we select several datasets including Arrhythmia~\citep{647926}, Netsim~\citep{smith2011network}, CMNIST-75sp~\citep{fan2022debiasing}, Causalogue~\citep{chen2024causal}, and baseline approaches including PC, SGS~\citep{kalisch2007estimating}, ACD~\citep{lowe2022amortized}, AVICI~\citep{lorch2022amortizedinferencecausalstructure}, DAG-GNN~\citep{yu2019dag}, CVAE~\citep{alfakih2023deep}, CAE~\citep{chen2023affective}, biCD~\citep{chen2023learning} for four combinations of data types (single-structure data + simple variables, single-structure data + complex variables, multi-structure data + simple variables, multi-structure data + complex variables), as shown in Table~\ref{tabdifferentdatatype}. For each method, we extend it as much as possible to the types of datasets that can be extended. Table~\ref{tabinconsistency} shows their performance on structure estimation, representation generation, and causal inconsistency.

Table~\ref{tabinconsistency} demonstrates that indefinite data (multi-structure data + complex variables, reflected in Causalogue) exhibits unique and significant causal inconsistencies, accompanied by equally poor performance estimating causal structure and representation. However, such significant ineffectiveness does not exist when multi-structure data or complex variables are present alone (see columns in Netsim and CMNIST-75sp). It appears that the simultaneous presence of multi-structure data and complex variables leads to the emergence of ineffectiveness. 

In conjunction with the two reconstruction losses discussed in Section~\ref{secindefinitedatacausaldiscovery} (loss 1=$distance(\hat{A}^{s},A)$, loss 2=$distance(\hat{A}^{r},A))$, it appears that the significant inconsistency between $\hat{A}^{r}$ and $\hat{A}^{s}$ arises due to the independent backpropagation from $\hat{A}^{s}$ and $\hat{A}^{r}$ initiated by the two reconstruction losses. Consequently, we attempt to introduce an additional loss to measure the distance between $\hat{A}^{r}$ and $\hat{A}^{s}$ (i.e., $distance(\hat{A}^{s}, \hat{A}^{r})$) during training. However, under the practical results under different models and distance functions, this loss function does not significantly reduce inconsistency. We provide a detailed account of the experimental implementation and results in Appendix~\ref{secfollowupexperimentsonconsistency}, which indicates that the training signal provided by $distance(\hat{A}^{s}, \hat{A}^{r})$ is far from sufficient to optimize the causal inconsistency.

\section{SSL Framework}
\label{secmethod}
\begin{figure*}
  \includegraphics[width=1\textwidth]{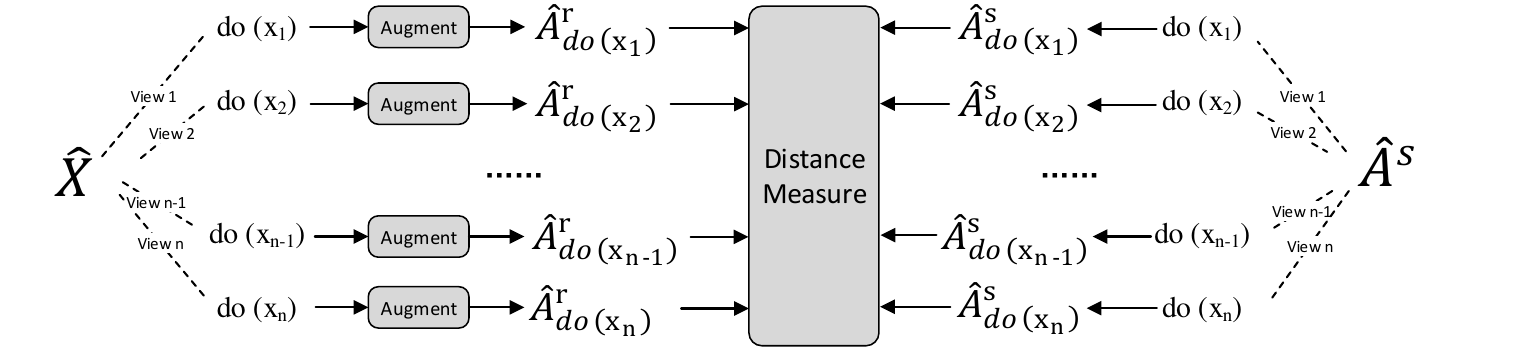}
  \caption{The SSL framework for causal consistency. The grey 
  rectangular boxes represent modules that require 
  specific implementation. From left to center, the process 
  describes how causal representations $\hat{X}$ are transformed into 
  the intervened causal structure $\hat{A}^{r}$. From right to center, the process 
  illustrates how causal structures $\hat{A}^{s} (\mathcal{G})$ are converted into intervened causal structure $\hat{A}^{s}$.}
  \label{figssl}
\end{figure*}
\subsection{Formulation Architecture} 
In order to excavate a more informative causal relationship between $\hat{A}^{s}$ and $\hat{A}^{r}$, we design an intervention-based self-supervised learning (SSL) framework. Broadly speaking, we attempt to apply various interventions $do(x_{1})$, $do(x_{2})$, etc. to $\hat{X}$, resulting in the intervened adjacency matrices $\hat{A}^{r}_{do(x_{1})}$, $\hat{A}^{r}_{do(x_{2})}$, etc. Concurrently, corresponding interventions are performed on $\hat{A}^{S}$ to obtain $\hat{A}^{s}_{do(x_{1})}$, $\hat{A}^{s}_{do(x_{2})}$, etc. According to the principle of causal abstraction~\citep{beckers2019abstracting} (refer to Appendix~\ref{suppproof} to see detailed proves), if $\hat{A}^{r}_{do(x_{1})}$, $\hat{A}^{r}_{do(x_{2})}$, ... correspondingly equal $\hat{A}^{s}_{do(x_{1})}$, $\hat{A}^{s}_{do(x_{2})}$, ..., the causal model of $\hat{X}$ and $\hat{A}^{s}$ is also equivalent. We elucidate the key roles of SSL within our framework as follows: 
\begin{itemize}
    \item \textbf{View}: We regard the intervention on different variables as a ``view". For instance, $do(x_{1})$ is considered a view. Specific operation of intervention is given in Section~\ref{secintervention}.

\item \textbf{Augmentation}: We refer to the specific operation of obtaining $\hat{A}^{r}_{do(x_{n})}$ from $\hat{X}$ as ``augmentation"\footnote{Augmentation does not exist in the process of obtaining $\hat{A}^{s}_{do(x_{n})}$ from $\hat{A}^{s}$, as the intervention on the structures can directly yield the result.}.

\item \textbf{Philosophy}: We define the distance measure corresponding to $\hat{A}^{s}_{do(x_{1})}$, $\hat{A}^{s}_{do(x_{2})}$,... and $\hat{A}^{r}_{do(x_{1})}$, $\hat{A}^{r}_{do(x_{2})}$,... as ``philosophy", describing that under the sufficient intervention set of two models, if the relationship strengths in the intervening adjacency matrices are equal, then the two models should also be equal.
\end{itemize}

Overall, as explained in Figure~\ref{figssl}, $\hat{A}^{r}_{do(x_{1})}$, $\hat{A}^{r}_{do(x_{2})}$,... and $\hat{A}^{s}_{do(x_{1})}$, $\hat{A}^{s}_{do(x_{2})}$,... originate from different intervention views of $\hat{X}$ and $\hat{A}^{s}$, representing adjacency matrices that reflect causal relationships under corresponding interventions. Augmentation signifies the specific measures to transform causal representations into causal relationships under corresponding interventions (for instance, the previously mentioned causal classifier $f_{c}$). Meanwhile, the distance measure of adjacency matrices under corresponding views represents the causal consistency under the respective interventions.
\subsection{Intervention}\label{secintervention}
Initially, we recall the conventional definition of intervention: 

\begin{equation}
   do(x_{t}=t):=P(x_{t}=t)=1
\end{equation}
where $t$ is one of the state which probably exists in the original distribution of variable $x_{t}$.
Interventions are typically represented by the $do$ operator, with the objective of setting the probability of an observed variable equaling a particular state to $1$. However, indefinite data considers complex variables, and causal representations exist in the form of high-dimensional continuous representations, leading to an unknown distribution. Hence, we propose the ``General Intervention'': 

\begin{proposition}[General Intervention]
General intervention is represented by the $do_{g}$ operator 
with the objective of setting the parent set 
of the observed variable to $\emptyset$. 

\begin{equation}
   do_{g}(x_{t}):=Pa(x_{t})=\emptyset
\end{equation}
where $Pa(x)$ represents a parent set of $x$ in the causal graph. 
\label{defgd}
\end{proposition}

The effect of $do_{g}(x_{t})$ 
is practically equivalent to the effect of the set of conventional interventions: 
$\{do(x_{t}=t_1),do(x_{t}=t_2),\dots\}$. 
%(For simplicity, unless specially stated, the term $do$ in the 
%rest of this paper represents either $do$ or $do_{g}$.)  
Specifically, ``General Intervention'' is a simplification of conventional intervention. When disregarding specific distributions, General Intervention can be viewed as treating all variables equivalently as binary variables ($do_{g}(x)$ seems like $do(x=0)=1$, and ``not $do_{g}(x)$'' seems like $do(x=1)=1$). Therefore, General Intervention is not a perfect intervention. 
\begin{figure*}
  \centering
  \subfigure[Causal Model $\mathcal{M}$]{
    \includegraphics[width=0.2\textwidth]{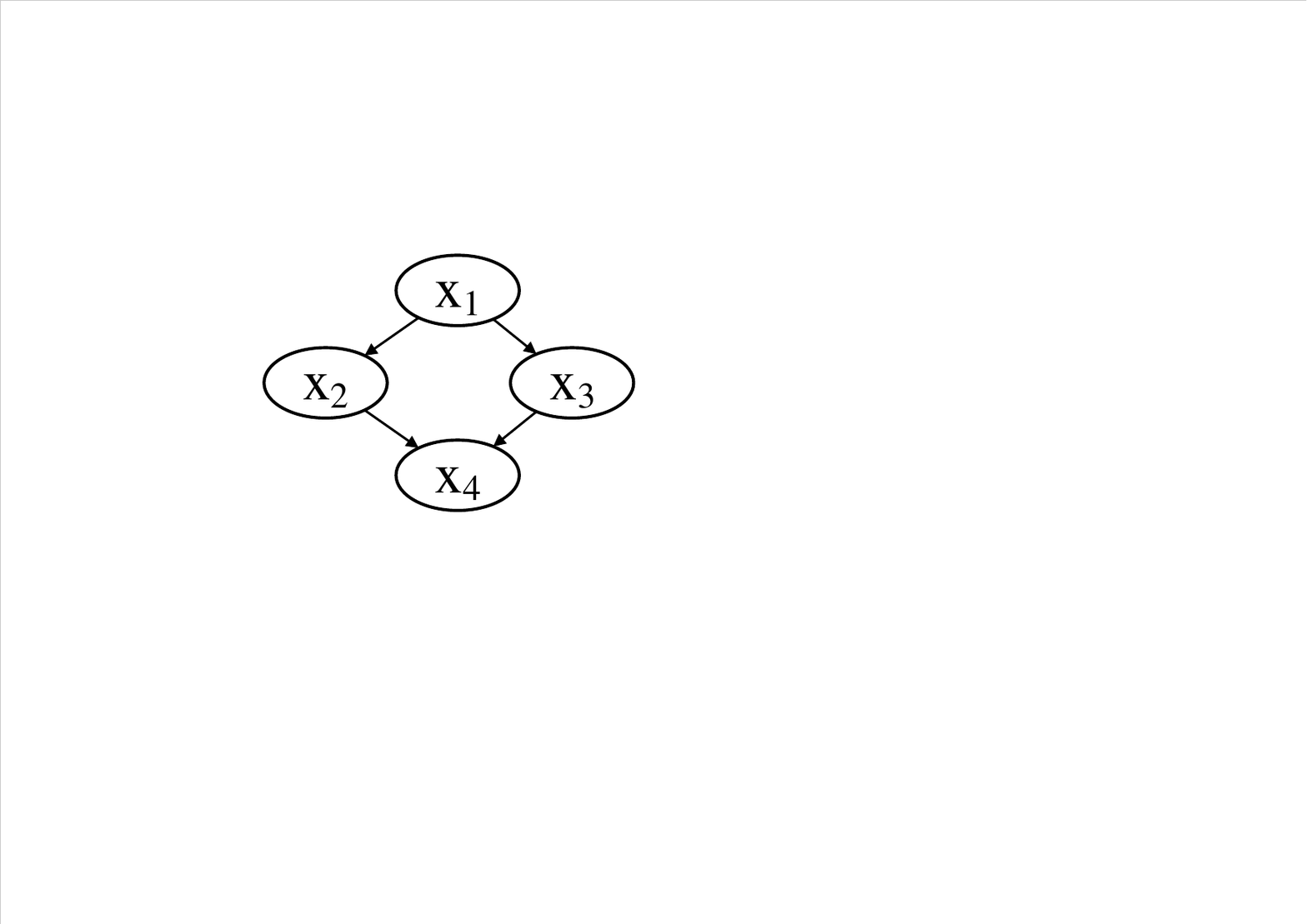}}
  \subfigure[Causal Model $\mathcal{N}$]{
    \includegraphics[width=0.2\textwidth]{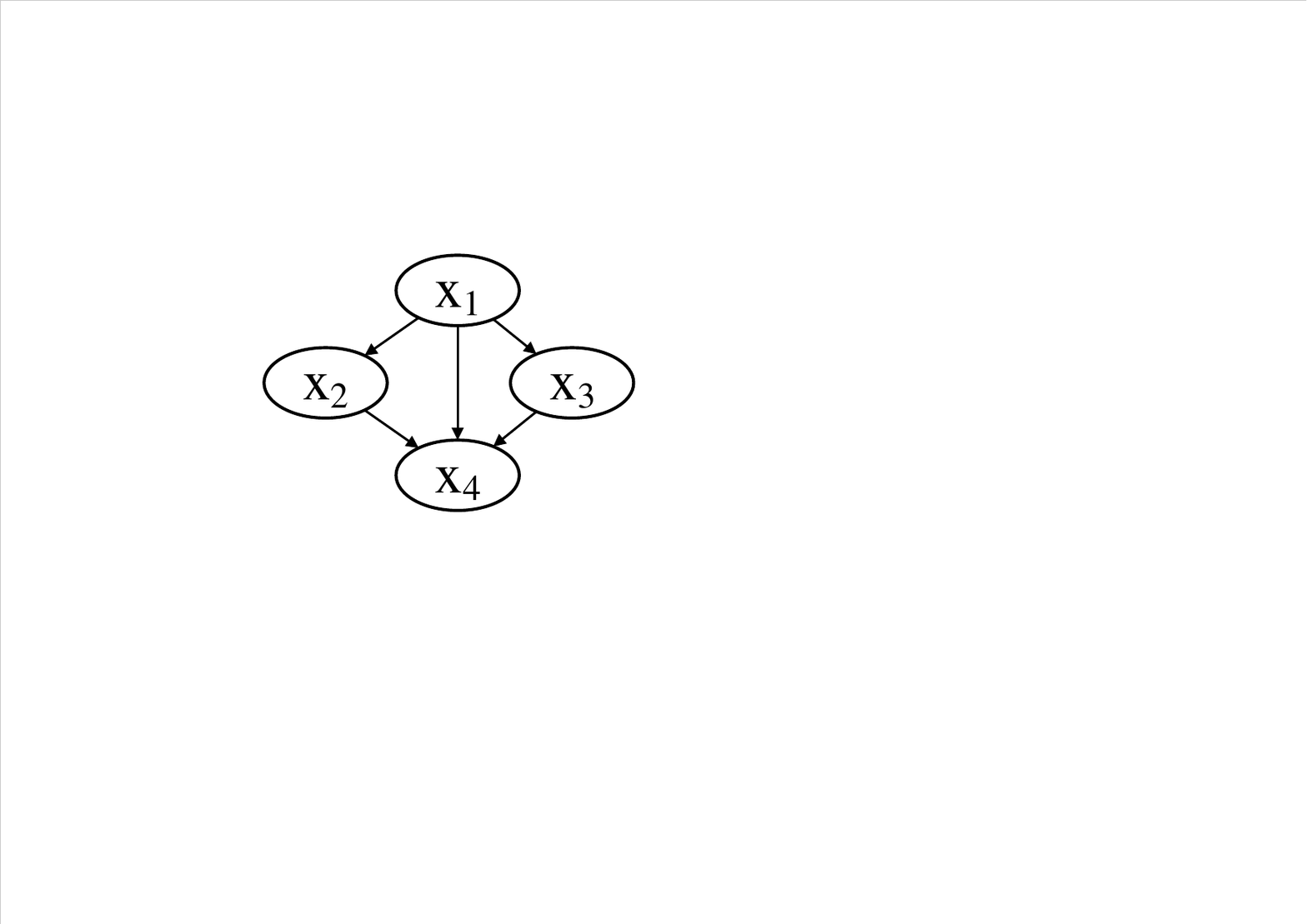}}
  
  \caption{Two cases of causal models having consistent relationships in diverse ``General Intervention''.}
  \label{figperfectintervention}
\end{figure*}
As shown in Figure~\ref{figperfectintervention}, in conventional interventions, two causal models $\mathcal{M}$ and $\mathcal{N}$ can observe whether $x_{1} \rightarrow x_{4}$ exists by intervening $x_{1}$ into different states ($t_1, t_2, ...$) in the distribution. However, in ``General Intervention'', $do_{g}(x_{1})$, $do_{g}(x_{2})$, $do_{g}(x_{3})$, and $do_{g}(x_{4})$ all yield completely identical sets of causal relationships between $\mathcal{M}$ and $\mathcal{N}$, because we can't observe the truth causal graph. When we conduct $do_{g}(x_{1})$, from model $\mathcal{M}$ we can get: $f(x_{1}, x_{2})=1$, $f(x_{1}, x_{3})=1$, $f(x_{1}, x_{4})=1$, and the same relationships can be got from model $\mathcal{N}$. Hence $do_{g}(x_{1})$ can't distinguish them.  thereby making it impossible to distinguish them. Naturally, we consider increasing the number of variables in each intervention, which can enhance the total number of interventions and thereby reveal more detailed causal relationships. For instance, a bivariate intervention can be written as: 

\begin{equation}
   do_{g}(x_{i}, x_{j}):=Pa(x_{i})=\emptyset~and~Pa(x_{j})=\emptyset
\end{equation} 
Theoretically, interventions involving more variables can provide more intervention views, thereby approaching a perfect intervention set. However, in practice, the most suitable number of intervention variables is related to the number of causal variables and the sparsity of the causal relationships. We demonstrate the impact of interventions on different variables on the results in our experiments (Figure~\ref{figinterpercent}).

\subsection{Optimization}
Assuming there exist $K$ intervention views, the distance measure function is $\mathcal{L}()$, the common learnable parameters of the model are $\theta$, and for any view $k$ ($1\leq k\leq K$), the learnable parameters of its augmentation are $\delta_{k}$. Then, given a dataset $\mathcal{X}$,we optimize: 

\begin{equation}
\begin{aligned}
    &\theta^{*}, \delta_{1}^{*},..., \delta_{K}^{*}= argmin_{\theta, \delta_{1},..., \delta_{K}}\mathcal{L}(\mathcal{X},\theta,\delta_{1},..., \delta_{K})\\
    &where: \mathcal{L}(\mathcal{X},\theta, \delta_{1},..., \delta_{K})=\sum_{k=1}^{K} l(\hat{A}^{r}_{do_{g}(x_{k})},\hat{A}^{s}_{do_{g}(x_{k})})
\end{aligned}
    \label{eqtpareto}
\end{equation}
Each view aims to optimize its objective such that its own structure and representation are consistent, thus multiple views collectively form a multi-objective optimization. 
Given Pareto optimality~\citep{censor1977pareto}, deleting $\delta_{1},..., \delta_{K}$ cannot make $l(\hat{A}^{r}_{do_{g}(x_{k})},\hat{A}^{s}_{do_{g}(x_{k})})$ of each view be optimal simultaneously (i.e., $\mathcal{L}(\mathcal{X},\theta)=\sum_{k=1}^{K} l(\hat{A}^{r}_{do_{g}(x_{k})},\hat{A}^{s}_{do_{g}(x_{k})})$ is impossible to find a parameter $\theta^{*}$ that minimizes the distance for every view). Therefore, we suggest that augmentation must include individual learnable parameters.

\subsection{Example for Implementation}
Taking an example from the variational probabilistic framework 
mentioned in Figure~\ref{figindefinitecausalmodel}, which has become a popular choice, 
%we simplify the encoder $p(z_{r}|X)$ as $p^{r}_{\varphi^{r}}$ with parameter $\varphi^{r}$, and the decoder $q(\hat{X}|z_{r})$ as $q^{r}_{\vartheta^{r}}$ with parameter $\vartheta^{r}$. Similarly, we simplify the encoder $p(z_{s}|(x_{i},x_{j}))$ as $p^{s}_{\varphi^{s}}$ with parameter $\varphi^{s}$, and the decoder $q(\hat{\mathcal{E}}_{i,j}|z_{s})$ as $q^{s}_{\vartheta^{s}}$ with parameter $\vartheta^{s}$. Hence, the causal representation $\hat{X}$ is generated by $p^{r}_{\varphi^{r}}$ and $q^{r}_{\vartheta^{r}}$, and the causal structure $\hat{\mathcal{E}}_{i,j}$ is generated by $p^{s}_{\varphi^{s}}$ and $q^{s}_{\vartheta^{s}}$. 
let adjacency matrix $\hat{A}^{r}$ and $\hat{A}^{s}$ represent the causal relationships of $\hat{X}$ and $\hat{\mathcal{E}}_{i,j}$, separately. For example, 
$\hat{A}^{s}_{i,j} \in [0,1]$ describes the strength of 
how variable $x_{j}$ influences 
variable $x_{i}$ from causal structure $\hat{\mathcal{E}}_{i,j}$, and $\hat{A}^{r}_{i,j}$ describes the causal relationships from causal representation $\hat{X}$. 

For the intervention $do_{g}(x_{k})$, we directly set the i-th row of $\hat{A}^{s}$ to zero, then $\hat{A}^{s}_{do_{g}(x_{k})}$ can be written as: 

\begin{equation}
\begin{split}
     (\hat{A}^{s}_{do_{g}(x_{k})})_{m,n}=\left\{
       \begin{array}{lr}
         0, &(m=i)\\
         \hat{A}^{s}_{m,n}, &else
       \end{array}
     \right.
   \end{split}
\end{equation}

Naturally, $\hat{X}_{do_{g}(x_{k})}=(I-\hat{A}^{s}_{do_{g}(x_{k})})^{-1}E$. As mentioned in Equation~\ref{eqtpareto}, we hope that the augmentation possesses independent learnable parameters, so we use an individual $MLP_{do_{g}(x_{k})}$ to convert causal representations into causal relationships (i.e., the previously mentioned causal classifier $f_{c}$). Specifically, for any two variables $x_{m}$ and $x_{n}$ in $\hat{X}_{do_{g}(x_{k})}$, 
\begin{equation}
   (\hat{A}^{r}_{do_{g}(x_{k})})_{m,n}=\sigma(MLP_{do_{g}(x_{k})}([x_{m},x_{n}]))
\end{equation}
where $\sigma(\cdot)$ is a sigmoid function projecting the output of MLP into the range of (0,1), which indicates the probability (causal relationship) of edge (m,n) under the intervention $do_{g}(x_{k})$. 

Finally, the distance measure $\mathcal{L}$ is set as MSE function: 
\begin{equation}
   \mathcal{L}=\sum^{K}_{k=1}MSE(\hat{A}^{r}_{do_{g}(x_{k})},\hat{A}^{s}_{do_{g}(x_{k})})
\end{equation}

\section{Experiments on SSL Framework}
\label{secexperiments}

\subsection{Datasets, Baselines, and Metrics} 
We evaluate our method on 2 Indefinite Datasets: 
\textit{Causalogue} and \textit{Causaction}~\citep{chen2024causal}.
The Indefinite Datasets are for Causal Discovery in Indefinite Data (CDID) task (producing the causal structures and causal representations as discussed in Section~\ref{secindefinitedatacausaldiscovery}), contributing to the 
\textbf{Main Results}. 
The experiments also incorporate a variety of baselines, encompassing causal deep models such as 
ACD~\citep{lowe2022amortized},  
DAG-GNN~\citep{yu2019dag}, 
ACCD~\citep{chen2023affective}, 
biCD~\citep{chen2023learning}, and intervention deep models 
like DisC~\citep{fan2022debiasing}, 
DIR~\citep{wu2022discovering}, and our method ($Ours$). We provide the details about all datasets, baselines and implementation in Appendix~\ref{suppddbi}.  

\subsection{Main Results}
We conduct experiments for CDID (Causal discovery in indefinite data) task and causal consistency on the \textit{Causalogue} and \textit{Causaction} datasets. 
CDID task is evaluated through the accuracy 
of causal structures and causal representations, respectively. 
Causal consistency is assessed by measuring the distance 
between $\hat{A}^{s}$ and $\hat{A}^{r}$. As demonstrated by Table~\ref{tabrssmcd}, 
our method significantly improves causal consistency, 
which correspondingly leads to an enhancement 
of causal accuracy, even compared with the SOTA model (biCD) for Indefinite data. 
This also indicates the fact that, until now, the causal 
consistency in Indefinite Data has often been overlooked though it is a crucial problem. 

\begin{table*}
  \centering
  \resizebox{1\textwidth}{!}{
  \begin{tabular}{c|cccccc|cccccc}
    \hline
    \bf Methods&\multicolumn{6}{c}{\textbf{Causalogue}} \vline&\multicolumn{6}{c}{\textbf{Causaction}}\\
    &\multicolumn{2}{c}{\textbf{Structure}}&\multicolumn{2}{c}{\textbf{Representation}}&\multicolumn{2}{c}{\textbf{Consistency}}\vline&\multicolumn{2}{c}{\textbf{Structure}}&\multicolumn{2}{c}{\textbf{Representation}}&\multicolumn{2}{c}{\textbf{Consistency}}\\
    &AUROC$\uparrow $&HD$\downarrow $&AUROC$\uparrow $&F1$\uparrow $&AUROC$\uparrow $&1-MSE$\uparrow $&AUROC$\uparrow $&HD$\downarrow $&AUROC$\uparrow $&F1$\uparrow $&AUROC$\uparrow $&1-MSE$\uparrow $\\
    \hline
    ACD         &0.55     &0.82$_{\pm 0.01}$     &0.55     &0.88     &0.51$_{\pm 0.01}$     &0.49$_{\pm 0.11}$     &0.65     &0.75$_{\pm 0.03}$      &0.59       &0.69       &0.55$_{\pm 0.09}$       &0.52$_{\pm 0.07}$\\
    DAG-GNN     &0.41     &0.78$_{\pm 0.04}$     &0.50     &0.88     &0.50$_{\pm 0.01}$     &0.49$_{\pm 0.12}$     &0.59     &0.71$_{\pm 0.05}$      &0.63       &0.64       &0.53$_{\pm 0.15}$       &0.51$_{\pm 0.12}$\\
    DisC    &0.58     &0.66$_{\pm 0.02}$     &0.58     &0.87     &0.52$_{\pm 0.00}$     &0.60$_{\pm 0.08}$     &0.65     &0.66$_{\pm 0.02}$      &0.66       &0.71       &0.62$_{\pm 0.11}$       &0.66$_{\pm 0.08}$\\
    DIR     &0.57     &0.68$_{\pm 0.06}$     &0.59     &0.86     &0.51$_{\pm 0.01}$     &0.63$_{\pm 0.13}$     &0.67     &0.61$_{\pm 0.05}$      &0.67       &0.70       &0.61$_{\pm 0.08}$       &0.71$_{\pm 0.12}$\\
    ACCD        &0.46     &1.02$_{\pm 0.05}$     &0.63     &0.92     &0.60$_{\pm 0.05}$     &0.69$_{\pm 0.11}$     &0.72     &0.83$_{\pm 0.07}$      &0.70       &0.81       &0.64$_{\pm 0.07}$       &0.77$_{\pm 0.05}$\\
    biCD        &0.66     &0.56$_{\pm 0.10}$     &0.65     &0.89     &0.89$_{\pm 0.04}$     &0.80$_{\pm 0.07}$     &0.78     &0.49$_{\pm 0.08}$      &0.75       &0.85       &0.79$_{\pm 0.04}$       &0.84$_{\pm 0.06}$\\
    \hline
    Ours&\bf 0.69 &\bf 0.49$_{\pm 0.05}$ &\bf 0.67 &\bf 0.95 &\bf 0.95$_{\pm 0.01}$ &\bf 0.95$_{\pm 0.07}$ &\bf0.79  &\bf0.36$_{\pm 0.05}$   &\bf0.78    &\bf0.96    &\bf0.89$_{\pm 0.05}$    &\bf0.87$_{\pm 0.04}$\\
    \hline
  \end{tabular}}
  \caption{Results on \textit{Causalogue} and \textit{Causaction} Datasets with 
  95$\%$ confidence interval shown.
  All evaluation metrics (except HD) are normalized to 
 the range $[0,1]$. Those $std$ of the results are omitted when $std<0.01$.}
  \label{tabrssmcd}
\end{table*}

Our findings also reveal some additional conclusions. 
As shown among intervention-related baselines, 
the intervention methods proposed by DisC and DIR could 
enhance the causal structure identification capability. 
This improvement is attributed to that interventions can adapt models to cross multi-structure environment. 
However, their interventions introduce bias that 
lies in forming negative samples by combining the causal pattern 
with the background from other samples in the batch, 
while ours does not adopt negative samples (by setting it as $\emptyset$). 
Their intervention concepts do enhance the 
model's discriminative capacity for causal patterns and shortcuts,  
but the sparsity of Indefinite Data samples has been 
introduced as bias into representation learning. 
In addition, ACD and biCD are methods specifically targeted 
at multi-structure data and complex variables, respectively. 
Therefore, their advantages have been particularly emphasized in our 
experimental results.

\begin{figure*}
  \centering
  \subfigure[Intervention Percentage (\%)]{
    \includegraphics[width=0.45\textwidth]{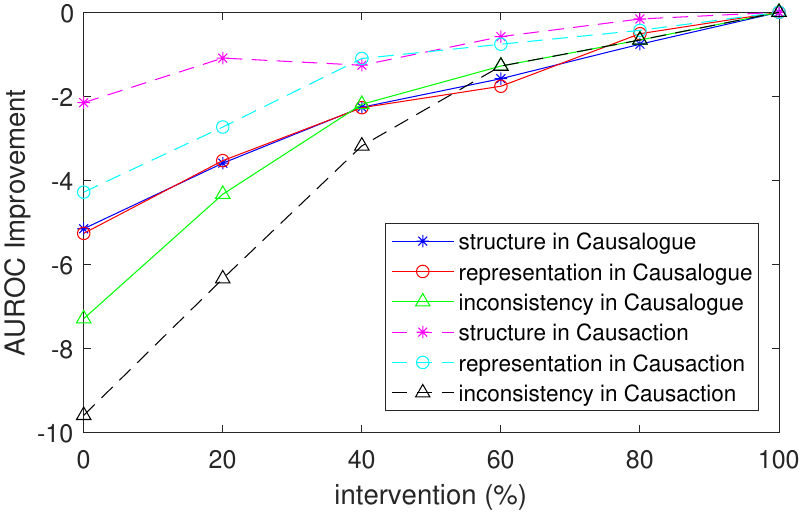}}
  \subfigure[\# Intervention Variables]{
    \includegraphics[width=0.45\textwidth]{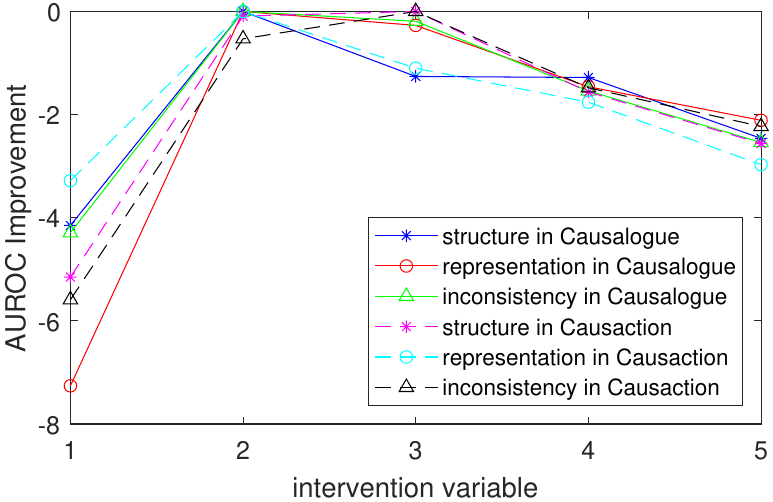}}
  \caption{Performance of the Ours under different sizes and different variable numbers 
  of intervention. ``Intervention 20$\%$'' 
  refers to an intervention set composed by randomly selecting 
  20$\%$ $do_{g}$ operators from the maximum intervention set.}
  \label{figinterpercent}
\end{figure*}

To evaluate the effect of the interventions under different levels to results, 
we test the performance of our method in \textit{Causalogue} dataset, under conditions ranging from an 
empty intervention set (no interventions carried out) 
to the maximum intervention set (all interventions carried out), when intervening two variables (Figure~\ref{figinterpercent} (a)). We also vary the number of intervening variables from 1 to 5 as shown in Figure~\ref{figinterpercent} (b).  

Figure~\ref{figinterpercent} demonstrates that increasing the number of intervening variables is not necessarily optimal. In the \textit{Causalogue} dataset, the best performance is achieved when there are two intervention variables. Moreover, as shown in Figure~\ref{figinterpercent} (a), continuously increasing the interventions does indeed enhance model performance, which corroborates our SSL view that ``the more views, the more comprehensive the information being capture".

Furthermore, we conduct an ablation study to determine the 
contribution of each mechanism. Table~\ref{tabablation} presents the following ablation studies:
(1) \textit{Augmentation}: The learnable parameters of augmentation for each view are set to be common; (2) \textit{cos\_sim}: The distance metric function is changed from MSE to cosine similarity; (3) \textit{Encoder}: The encoder for causal representation is changed from $(I-\hat{A}^{s})X$ to a learnable unknown distribution (implementation details in Appendix~\ref{secid}); (4) \textit{Decoder}: The decoder for causal representation is changed from $(I-\hat{A}^{s})^{-1}X$ to a learnable unknown distribution. 
Table~\ref{tabablation} confirms that common augmentation parameters lead to a decrease in performance, which corroborates our proposed Pareto solution. The replacement of the distance metric function has little impact on the model, as the adjacency matrix is already a low-dimensional representation of the causal model, and the differences under different measures are not significant. However, the replacement of the encoder and decoder results in the loss of the causal mechanism, leading to a significant performance decrease. This indicates that the causal mechanism underpins the discovery of the entire causal structure and causal representation. 

\begin{table}
  \centering
  \resizebox{0.6\textwidth}{!}{
  \begin{tabular}{c|ccc}
    \hline
    \bf Model&\bf Structure&\bf Representation&\bf Consistency\\
    \hline
    augmentation&$\downarrow$ 0.03&$\downarrow$ 0.02&$\downarrow$ 0.02\\
    cos\_sim&$\downarrow$ 0.01&$\downarrow$ 0.01&$\downarrow$ 0.00\\
    encoder&$\downarrow$ 0.05&$\downarrow$ 0.08&$\downarrow$ 0.07\\
    decoder&$\downarrow$ 0.04&$\downarrow$ 0.08&$\downarrow$ 0.06\\
    \hline
  \end{tabular}}
  \caption{Ablation Results of AUROC on three measures. }
  \label{tabablation}
\end{table}

Moreover, in Table~\ref{tabrssmcd}, we only show the evaluation results between 
structure and representation. To further demonstrate 
the benefits of our SSL framework to the model, 
we additionally focus on the error of the similarity matrices 
of structure to the ground truth, and representation to 
the ground truth, respectively. Figure~\ref{figvariance} 
shows the box plots of the errors in structure and representation. 
Ours evidently outperforms other methods. The details are 
shown in Appendix~\ref{expvlr}. Lastly, We evaluate scalability by scaling the training set. 
Table~\ref{tabscalability} shows that our method performs best 
under any scale of datasets, especially in terms of structure. 
The details are shown in Appendix~\ref{expscalability}.

\section{Application}
\label{secapplication}
\subsection{Downstream Tasks} 

\subsubsection{Task Definitions}
Firstly, we apply our SSL framework to several downstream tasks requiring causal structure and causal representation learning, namely Emotion Recognition in Conversation (ERC), Emotion-Cause Pair Extraction (ECPE), and Temporal Action Segmentation (TAS), of which two are text modality: ERC and ECPE, and one is video modality: TAS. The definitions of these tasks are as follows:

\noindent\textbf{ERC} task: Given a conversation $D=\{U_{1},..., U_{N}\}$ consisting of $N$ utterances, the output is the emotion label $Emo$ contained in each utterance, i.e., the output is $\{Emo_{1},...,Emo_{N}\}$.

\noindent\textbf{ECPE} task: Given a text $D=\{S_{1},..., S_{N}\}$ consisting of $N$ sentences, the output is all pairs $(S_{i}, S_{j})$ that satisfy the condition where $S_{i}$ expresses an emotion and $S_{j}$ is the cause of the emotion in $S_{i}$.

\noindent\textbf{TAS} task: Given a video $V$, the output is several segments $\{Seg_{1},..., Seg_{N}\}$. The sum of the frames of these segments is equal to the number of frames in video $V$, i.e., $Seg_{1}+Seg_{2}+...+Seg_{N}=V$. Each segment represents an action, so the action label corresponding to each segment, i.e., $\{Act_{1},..., Act_{N}\}$, needs to be produced.

Existing works for these three tasks emphasize the importance of causal analysis~\citep{shen-etal-2021-directed,chen2023affective,du2023casr}. Specifically, the ERC task requires causal structure learning to obtain the causal relationships between each utterance, and then decides which contexts to consider for predicting emotions through causal representation learning. The ECPE task requires causal structure learning for each sentence to obtain the causal relationships, and uses causal representation learning to predict the corresponding emotions and determine whether the representation pair of any two sentences meets the task requirements for the \textit{(Emotion, Cause)} pair. The TAS task requires causal structure learning to obtain the causal relationships between actions and feedback to the action labels, and it also requires causal representation learning to determine which frame-level representation is closer to the segment-level representation.  

\subsubsection{Datasets, Metrics, and Implementations} 

\begin{table*}
  \centering
  \resizebox{1\textwidth}{!}{
  \begin{tabular}{c|cccc}
    \hline
    \bf Ours&\bf Tasks&\bf Datasets&\bf Baselines&\bf Metrics\\
    \hline
    \multirow{3}{*}{Tasks}&ECPE&RECCON&ACCD, biCD, EDKA-GM, seF&F1\\
    &ERC&MELD, EmoryNLP, DD, IEM&ACCD, biCD, DAG-ERC, DualGAT, MultiEMO&F1\\
    &TAS&GTEA, 50salads, Breakfast&MS-TCN++, ASRF, CETNet, C2F&acc, Edit, F1$@$k,C-Dis\\
    \hline
    LLM&CSL&\textit{Causalogue}, RECCON&Zero-shot, Zero-shot-Cot, Auto-Cot& F1\\
    \hline
  \end{tabular}}
  \caption{Summarization of datasets, baselines and metrics}
  \label{tabsdbm}
\end{table*}

We evaluate these 3 tasks on 8 datasets: RECCON~\citep{poria2021recognizing}, 
MELD~\citep{poria-etal-2019-meld}, 
EmoryNLP~\citep{zahiri:18a}, 
DD~\citep{li-etal-2017-dailydialog}, 
IEM~\citep{busso2008iemocap}, 
GTEA~\citep{fathi2011learning}, 
50salads~\citep{stein2013combining}, and 
Breakfast~\citep{kuehne2014language} and their corresponding baselines: such as EDKA-GM~\citep{li2023experiencer}, 
seF~\citep{li2023class} for ECPE task, 
DAG-ERC~\citep{shen-etal-2021-directed}, 
DualGAT~\citep{zhang2023dualgats}, 
MultiEMO~\citep{shi2023multiemo} for ERC task, 
and MS-TCN++~\citep{li2020ms}, 
ASRF~\citep{ishikawa2021alleviating}, 
CETNet~\citep{wang2023cross}, and 
C2F~\citep{singhania2021coarse} for TAS task.  
For different downstream tasks, 
we utilize their prevalent metrics for evaluations. 
The criss-cross relationships between these datasets, 
tasks, metrics, and baselines have been summarized  
in Table~\ref{tabsdbm} and the details 
of them are shown in Appendix~\ref{suppddbi}. Moreover, we treat the SSL framework as a plug-in module, enhancing the performance of the original backbone by conducting contrastive learning on the causal structure and representation output by the original backbone. For the ECPE task, we use biCD as the backbone model. For the ERC task, we use DualGAT as the backbone model, and for the TAS task, we use CETnet as the backbone model. different pre-training models and implementation 
parameters are provided in Appendix~\ref{secid}. 

\subsubsection{Quantitative results}
\begin{table}
   \centering
   \resizebox{0.85\textwidth}{!}{
   \begin{tabular}{cc|ccccc}
     \hline
     \multicolumn{2}{c}{\textbf{ECPE}}\vline &\multicolumn{5}{c}{\textbf{ERC}}\\
     \bf Model&\bf RECCON&\bf Model&\bf MELD&\bf EmoryNLP&\bf DD&\bf IEM\\
     \hline
     ACCD&73.17$_{\pm 1.1}$&ACCD&63.81$_{\pm 0.11}$&39.54$_{\pm 0.12}$&59.53$_{\pm 0.01}$&69.17$_{\pm 0.15}$\\
     biCD&74.14$_{\pm 0.74}$&biCD&63.22$_{\pm 0.17}$&38.21$_{\pm 0.11}$&59.64$_{\pm 0.07}$&67.15$_{\pm 0.09}$\\
     EDKA-GM&72.14$_{\pm 0.93}$&DAG-ERC&63.65$_{\pm 0.05}$&39.02$_{\pm 0.13}$&59.33$_{\pm 0.01}$&68.03$_{\pm 0.15}$\\
     seF&74.55$_{\pm 0.98}$&DualGAT& 66.72$_{\pm 0.12}$&40.88$_{\pm 0.15}$&61.80$_{\pm 0.02}$&67.74$_{\pm 0.21}$\\
     -&-&MultiEMO&61.23$_{\pm 1.26}$&37.14$_{\pm 0.11}$&57.46$_{\pm 0.01}$&64.41$_{\pm 0.16}$\\
     \hline
     Ours&\bf 76.89$_{\pm 1.21}$&Ours&\bf 67.79$_{\pm 0.18}$&\bf 40.95$_{\pm 0.08}$&\bf 62.57$_{\pm 0.01}$&\bf 69.81$_{\pm 0.26}$\\
     \hline
     
   \end{tabular}}
   \caption{Results on ECPE and ERC tasks with 
   95$\%$ confidence interval shown.
   The evaluation metric used in the table is F1 score. 
   The backbones of Ours are biCD and DualGAT, respectively.}
   \label{tabrssmee}
 \end{table}

 \begin{table}
  \centering
  \resizebox{1\textwidth}{!}{
  \begin{tabular}{c|cccccc|cccccc|cccccc}
    
    \hline
    \bf Model &\multicolumn{6}{c}{\textbf{GTEA}} \vline&\multicolumn{6}{c}{\textbf{50salads}} \vline&\multicolumn{6}{c}{\textbf{Breakfast}}\\
     &\multicolumn{3}{c}{F1$@\{10, 25, 50\}$}&Edit&Acc&C-Dis &\multicolumn{3}{c}{F1$@\{10, 25, 50\}$}&Edit&Acc&C-Dis&\multicolumn{3}{c}{F1$@\{10, 25, 50\}$}&Edit&Acc&C-Dis\\
    \hline
     MSTCN++&82.3&83.6&71.9&79.8&77.6&8.4&79.4&77.3&69.3&71.6&82.8&3.3&-&-&-&-&-&-\\
    ASRF&85.5&83.8&73.6&76.9&74.7&9.0&80.3&77.4&67.4&74.2&77.6&4.9&69.1&63.4&50.8&66.6&63.0&55.8\\
    CETnet&90.5&89.6&78.9&85.7&79.4&7.1&87.6&87.3&80.9&82.8&87.3&2.6&72.5&68.7&57&72.8&74.2&38.1\\
    C2F&88&86.6&78.3&81.6&\bf 80.6&7.4&83.5&81.5&71.8&75.7&86.9&2.8&71.6&68.0&57.1&68.1&74.6&49.8\\
    \hline
    Ours&\bf 91.4&\bf 90.2&\bf 80.5&\bf 87.2&79.7&\bf 6.9&\bf 88.9&\bf 87.6&\bf 81.4&\bf 83.1&\bf 88.9&\bf 2.5&\bf 78.7&\bf 74.9&\bf 63.4&\bf 78.3&\bf 75.6&\bf 35.4\\
    \hline
     
  \end{tabular}}
  \caption{Results on TAS task. 
  All evaluation metrics used in the table are shown in Table~\ref{tabsdbm} and introduced in Appendix~\ref{ssstas}. 
  The backbone of Ours is CETnet. }
  \label{tabrssmtas}
  \end{table}

We assess the performance of Our SSL framework (Ours)
on ECPE, ERC, and TAS. 
From the outcomes presented in Tables~\ref{tabrssmee} and~\ref{tabrssmtas}, Ours exhibits a clear improvement when comparing with these specific backbone models themselves. In the ECPE and ERC tasks, some baselines do not use causality-related methods, so we cannot quantitatively compare specific causal structures, causal representations, and causal consistency. However, in the TAS task, there are some similar common metrics: C-Dis represents the distance between action structure graphs, which can be used to approximately reflect differences in causal structures, while Edit represents the difference between frames, derived from frame-level representations, which can be used to measure the accuracy of causal representations. The underlying reason for enhancement is that, under the conditions of ensured causal consistency, an increase in the accuracy 
of the causal model promotes enhancements in both the causal structure (C-Dis in Table~\ref{tabrssmtas}) and causal representation (Edit in Table~\ref{tabrssmtas}), surpassing other methods, hence improving the final results. 

\subsubsection{Qualitative results}

\begin{figure}
   \centering
   \subfigure[Case1]{
     \includegraphics[width=\textwidth]{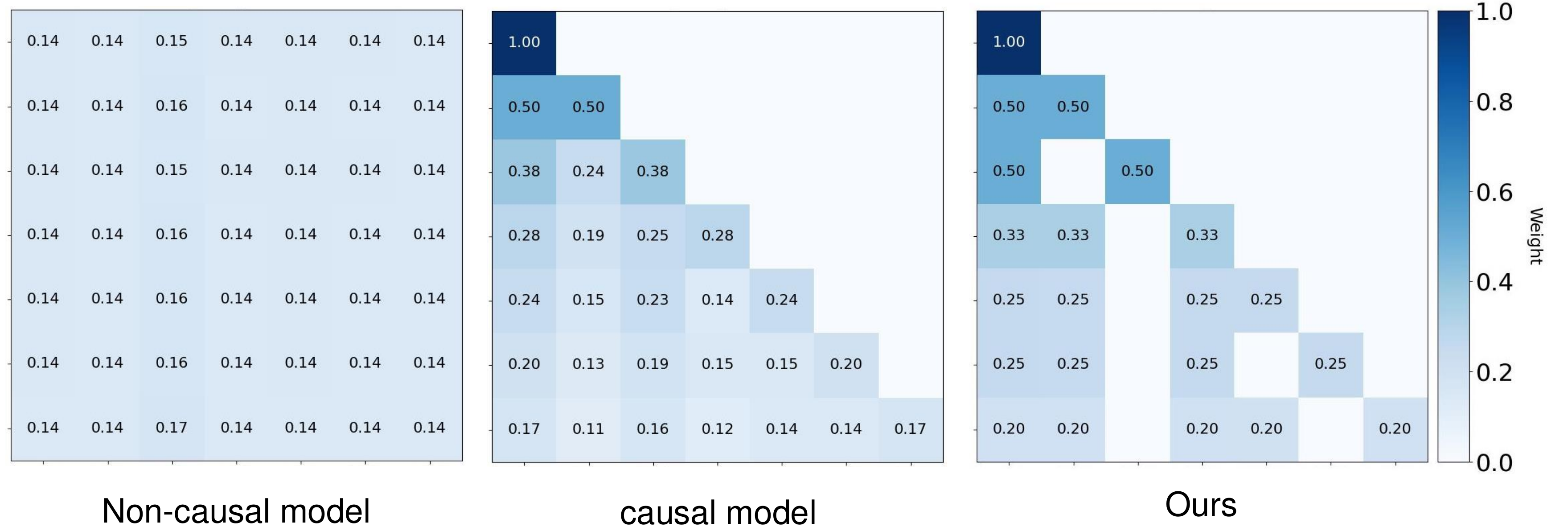}}
   \subfigure[Case2]{
     \includegraphics[width=\textwidth]{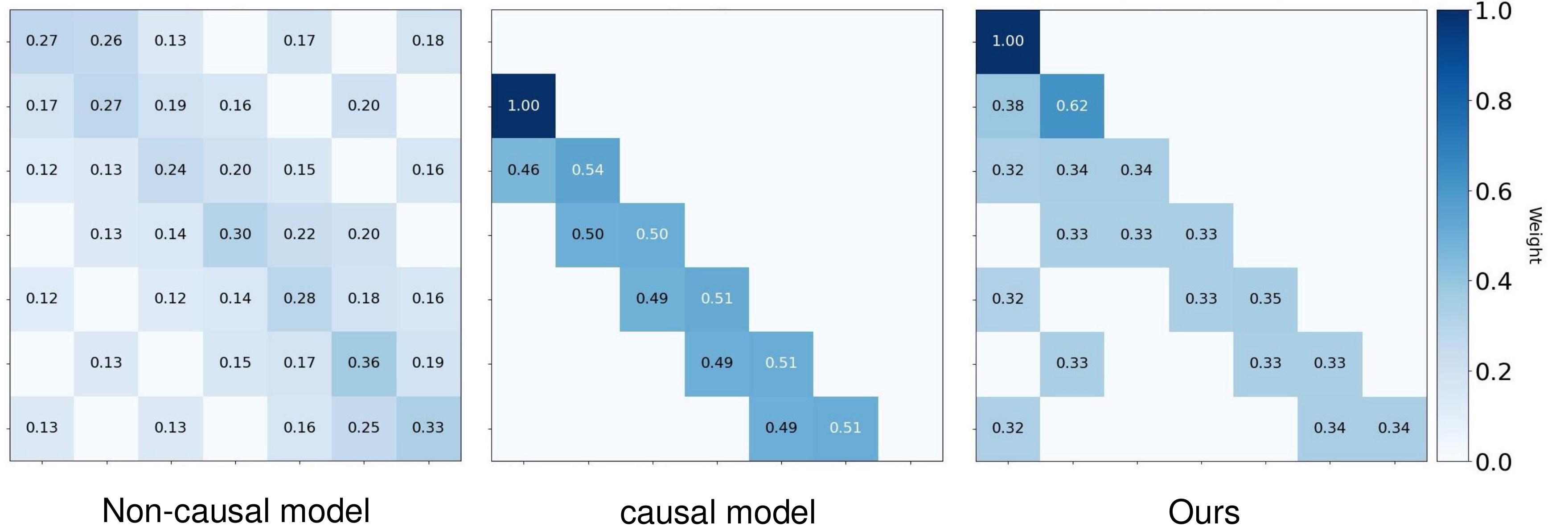}}
   \subfigure[Case3]{
     \includegraphics[width=\textwidth]{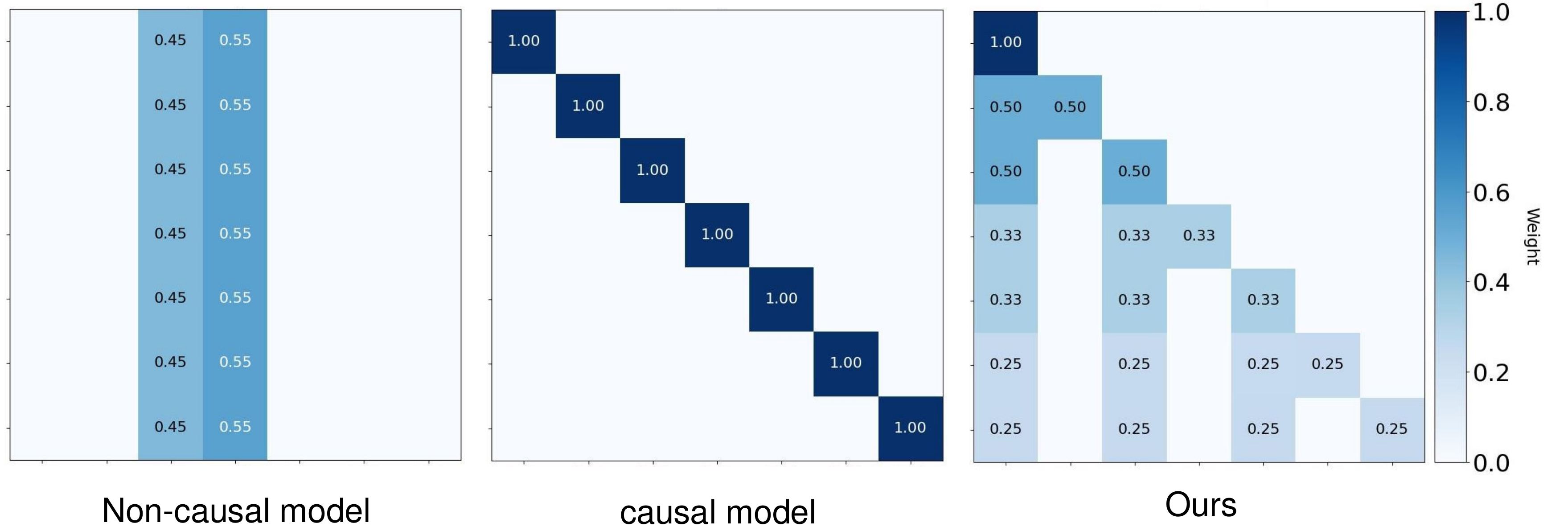}}
   
   \caption{Visualization of adjacency matrices of 3 cases on ECPE task. 
   We use EDKA-GM as the Non-causal model, and biCD as the causal model (the backbone model with our method in this task). 
   The adjacency matrix is $N\times N$, representing the relationship between any 
   two utterances.}
   \label{figvisualecpe}
 \end{figure}

\begin{figure}
   \centering
   \subfigure[Case1]{
     \includegraphics[width=\textwidth]{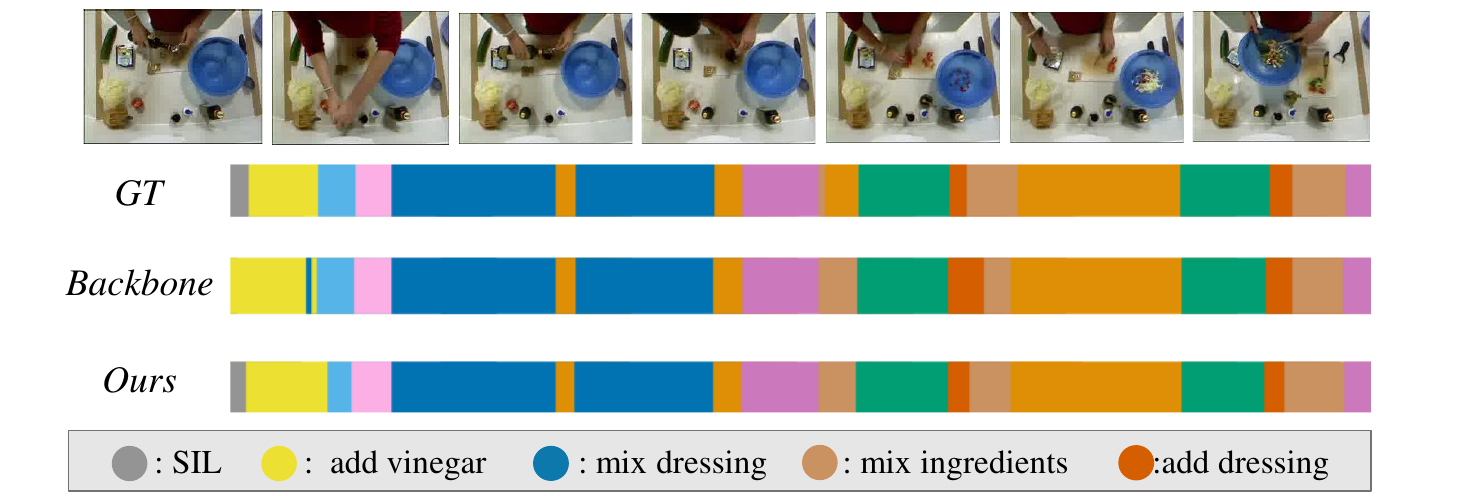}}
   \subfigure[Case2]{
     \includegraphics[width=\textwidth]{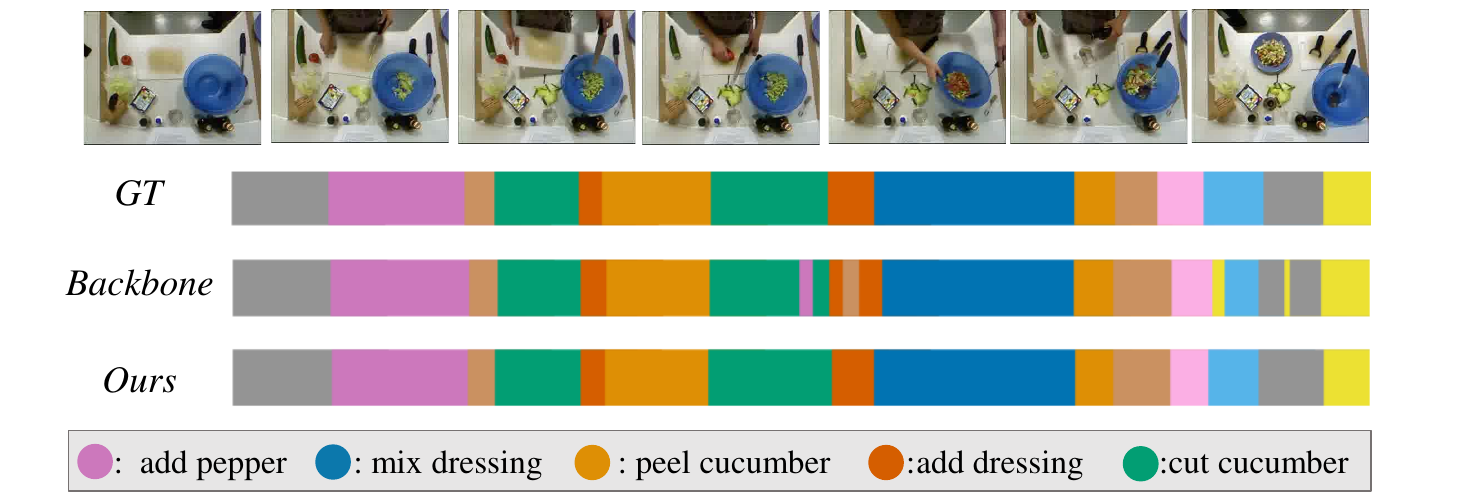}}
   \subfigure[Case3]{
     \includegraphics[width=\textwidth]{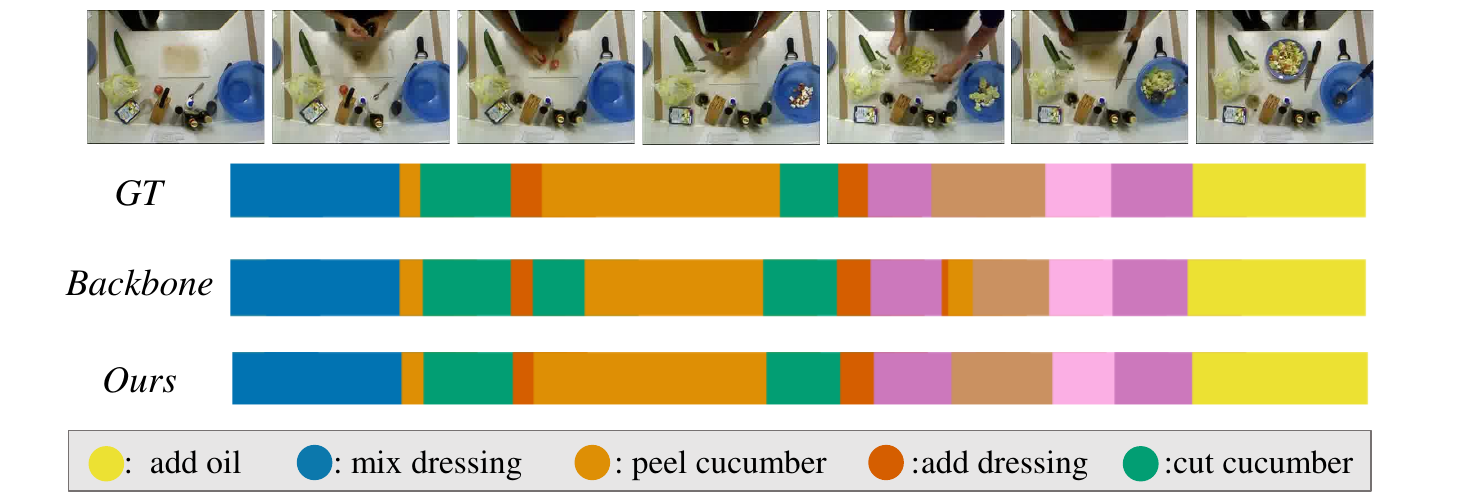}}

   \caption{Visualization of results of 3 cases on 50salads 
   dataset. GT represents the Ground Truth. The Backbone we choose is CETnet. }
   \label{figvisualtas}
 \end{figure}

To better illustrate the role of the causal model 
in these downstream tasks, we present two visualizations 
in Figures~\ref{figvisualecpe} and~\ref{figvisualtas}. 
Figure~\ref{figvisualecpe} displays a visualization 
of the adjacency matrix for the ECPE task, which can be 
equated with a causal graph, showing how the model assigns weights 
to the context when learning sentence relationships. 
Figure~\ref{figvisualecpe} demonstrates that the superiority of the causal method 
over non-causal ones lies in turning the adjacency matrix 
into a DAG, thus avoiding the factual error of treating earlier  
utterances as outcomes of latter ones. However, due to 
unknown causal labels, there is not a sufficiently 
strong constraint for causal graph, which often leaves 
the model uncertain about which edges in the DAG should exist. 
Our model mitigates this issue by using causal consistency 
constraints, enabling the model to identify the correct edges 
through contrastive learning of the causal representation 
and structure. 

Figure~\ref{figvisualtas} shows the TAS task's visualization results, 
illustrating that causal consistency between frames and segments  
can significantly reduce the existence of trivial segments. 
This harmonizes with Causal abstraction: 
all frames within a segment share a similar causal relationship. 
Moreover, after consistent learning of causal structures and representations, significant preemption is eliminated: for instance, ``peeling a cucumber'' cannot occur after ``cutting a cucumber''. The essence of these preemption stems from the lack of consistency with the causal structure at the representation level, leading to the inability of the causal relationship ``cutting a cucumber" $\rightarrow$ ``peeling a cucumber" to be effectively transferred from causal structure learning to causal representation learning.

\subsection{LLM inference} 
Another application focuses on the inference of Large Language Models (LLMs), specifically the inferences of causal relationships among sentences in the input text (i.e., a causal structure prediction task on some kind of indefinite data). Specifically, we treat each dialogue as a sample, 
where an utterance is regarded as a causal variable. That is, for a dialogue $D=\{Utt_1, Utt_2, \dots,Utt_N\}$, where $Utt_i$ represents $i$-th utterance, $N$ is the number of causal variables, the LLM should generate an answer indicating which pair of utterances possess causal relationships. 

It is well known that LLMs generate text based on the maximum likelihood of the next token, so their generalization ability on complex reasoning tasks is not sufficiently strong~\citep{vashishtha2023causal,zhang2024leveraging}. However, recent works based on instructions and Chain of Thought (CoT) have supported that providing sufficient information at the input stage can more effectively invoke the basic reasoning ability of LLMs~\citep{zhang2022automatic,kojima2022large}. Therefore, we also propose a ``self-supervised" iterative instruction. Simply put, each time the LLM predicts a causal relationship, the instruction will guide the LLM to judge the causal consistency of the text under different intervention perspectives of this causal relationship, and feedback those answers that cannot pass the causal consistency judgment to the LLM to regenerate the causal relationship. Specifically, there are 3 steps: 

\vspace{1mm}
\noindent\textbf{Step 1 (Prediction Causal Relationship)}: 
Given an input text $D=\{Utt_1, \dots,Utt_N\}$, LLMs are asked to generate a text $C$ to answer which pairs of utterances have causal relationship. The instruction employs input accompanied by a demonstration, as shown in the following:  
\begin{mdframed}[backgroundcolor=green!5]
\textit{``You are assuming the role of a researcher 
capable of distinguishing between causation and correlation, 
charged with the task of recognizing the causal relationships 
among individual utterances within a given dialogue. 
We prescribe that the judgment of causation between two utterances 
is based on whether the former is the intended target 
of the latter's response. Whereas, correlation is gauged on 
whether the two share similar topics or vocabulary. 
The following is an example:}
\end{mdframed}

\begin{mdframed}[backgroundcolor=green!5]
\textit{Dialogue:}

\textit{`1. Hazel drank too much champagne at the party.}

\textit{2. Oh my goodness! That sounds like quite an eventful party.}

\textit{3. Well, drinking too much alcohol can have many negative effects.}

\textit{4. Oh no, I can imagine Hazel waking up with a massive headache tomorrow.'}

\textit{Question 1: Is there a causal relationship from utterance 1 to 2?}

\textit{Answer 1: Yes.}

\textit{......}

\textit{Question 6: Is there a causal relationship from utterance 3 to 4?}

\textit{Answer 6: Yes.  }
\end{mdframed}

\begin{mdframed}[backgroundcolor=green!5]
\textit{Given the above example, with its associated questions and answers, consider the following dialogue:}

\textit{Dialogue:}

\textit{`1. Charlotte has no idea how to avoid massive estate taxes.}

\textit{2. Estate taxes are a topic of concern for people in various countries.}

\textit{3. So, does anyone else have any knowledge or ideas on how to reduce estate taxes?}

\textit{4. Oh, that reminds me of a story about my uncle.' }

\textit{Question 1: Is there a causal relationship from utterance 1 to utterance 2?}

\textit{......} 

\textit{Question 6: Is there a causal relationship from utterance 3 to utterance 4?"}
\end{mdframed}

Additionally, through exact matching of characters in the output text $C$, we can automatically construct the adjacency matrix $\hat{A}^{s}$ of causal relationships, where the elements are binary variables, with 0 indicating no causal relationship and 1 indicating the existence of a causal relationship.

\vspace{1mm}
\noindent\textbf{Step 2 (Intervention)}: 
For $\hat{A}^{s}$ obtained in the first step, we perform as many interventions as possible on it (default setting is bivariate intervention, i.e., $do_{g}(Utt_{i}, Utt_{j})$). For each adjacency matrix $\hat{A}^{s}_{do_{g}(Utt_{i},Utt_{j})}$ obtained from the intervention, we delete the utterances in the input text $D=\{Utt_1, Utt_2, \dots,Utt_N\}$ that correspond to the intervened parent nodes, re-input them, and obtain the intervened $\hat{A}^{r}_{do_{g}(Utt_{i},Utt_{j})}$ according to the instruction in Step 1.

\vspace{1mm}
\noindent\textbf{Step 3 (Inconsistency Feedback)}: 
We compare each pair of adjacency matrices $\hat{A}^{r}_{do_g}$ and $\hat{A}^{s}_{do_g}$ 
obtained from different views. If a condition occurs where 
$(\hat{A}^{s}_{do_g})_{i,j}=1$  while $(\hat{A}^{r}_{do_g})_{i,j}=0$, this implies that for the locally intervened inputs without parent nodes in intervention, there should not be a correlation between $Utt_{i}$ and $Utt_{j}$ (it could be $Utt_{i} \rightarrow Utt_{j}$ or a common cause arises between $Utt_{i}$ and $Utt_{j}$), while it still appears in $\hat{A}^{s}_{do_{g}}$ even though affected by the intervention. 
Hence, it can be inferred that \textit{``from the intervention, there is no common cause between 
the i-th utterance and the j-th utterance, 
and the i-th utterance should not 
have a causal relationship to the j-th utterance."} 
On the contrary, if a situation arises where $(\hat{A}^{s}_{do_g})_{i,j}=0$  while $(\hat{A}^{r}_{do_g})_{i,j}=1$, we can assert that \textit{
``from the intervention, there is a common cause between the i-th utterance and 
the j-th utterance, and the i-th utterance should  
have a causal relationship to the j-th utterance.''}
If '$i$' refers to the first utterance, no response will be given 
to the clause relevant to the \textit{`common cause.'}  
An example of feedback instruction is as follows: 

\begin{mdframed}[backgroundcolor=green!5]
\textit{``After intervention, there should be no common cause between the second utterance and the third utterance, and the second utterance should not have a causal relationship with the third utterance, and there should be no common cause between the third utterance and the fourth utterance, and the third utterance should not have a causal relationship with the fourth utterance. Please re-answer based on these circumstances.''}
\end{mdframed}

\textbf{Recursive Process}: 
The iterative algorithm is 
summarized in Algorithm~\ref{algori1}. 
Steps 1 to 3 will continuously iterate through. 
The end condition is reached once all instances of 
$\hat{A}^{r}_{do_g}$ and $\hat{A}^{s}_{do_g}$ across all views are identical. 
The causal relationship generated by the LLM during the final 
loop represents the final results. Step 2 represents 
the `Augment' module mentioned in Section~\ref{secmethod} 
while Step 3 embodies the `Consistency check'. 
The overall objective of the instruction is to enable 
the LLM to identify causal relationships between utterances 
without causal labels.  

\begin{algorithm}
  \caption{Iterative Instruction}
  \textbf{Require}: A dialogue text $D=\{Utt_1, Utt_2, \dots,Utt_N\}$, 
  a set of matrices $\hat{A}^{s}_{do_g}\_Set=\{\hat{A}^{s}_{do_g(x_i, x_j)}, i,j \in N ~and~ i\leqslant j\}=\emptyset $, 
  a set of matrices $\hat{A}^{r}_{do_g}\_Set=\{\hat{A}^{r}_{do_g(x_i, x_j)}, i,j \in N ~and~ i\leqslant j\}=\emptyset $, 
  and $input\_instrcution$ as shown in Step 1.
  
  \textbf{Ensure}: Causal relation adjacency matrix $\hat{A}^{s} \in \mathbb{R}^{N*N}$ 
  in where $\hat{A}^{s}_{do_g}\_Set$ is consistent with $\hat{A}^{r}_{do_g}\_Set$ (Both $\neq \emptyset $). 
  \begin{algorithmic}
  \State \textbf{procedure} INTERVENTION ($I_*=\{do_{i,j}\}$) ($i,j \in N ~and~ i\leqslant j$)
  \While{$\hat{A}^{s}_{do_g}\_Set=\hat{A}^{r}_{do_g}\_Set\neq \emptyset$}
  \State Predict $\hat{A}^{s}$ via LLM according to $input\_instrcution$. 
  \For{each view $do_{i,j}$ in $I_*$}
  \State Calculate $(\hat{A}^{s}_{do_g})_{i,j}$ via Step 1. 
  \State $\hat{A}^{s}_{do_g}\_Set=\hat{A}^{s}_{do_g}\_Set \cup (\hat{A}^{s}_{do_g})_{i,j}$ 
  \State Calculate $(\hat{A}^{r}_{do_g})_{i,j}$ via Step 2. 
  \State $\hat{A}^{r}_{do_g}\_Set=\hat{A}^{r}_{do_g}\_Set \cup (\hat{A}^{r}_{do_g})_{i,j}$
  \EndFor
  \State get the inconsistency and edit the $feedback\_instruction$
  \State Replace the $input\_instrcution$ with $feedback\_instruction$ 
  \State $\hat{A}^{s}_{do_g}\_Set=\hat{A}^{r}_{do_g}\_Set= \emptyset$
  \EndWhile
\end{algorithmic}
\label{algori1}
\end{algorithm}
Table~\ref{tabllm} 
illuminates some interesting conclusions on public GPT-4 using gpt-4-32k-0314~\citep{openai2023gpt4} (The details of experiments are shown in Appendix~\ref{suppddbi}) - existing instruction 
methods are difficult to yield effective outcomes for this task. 
For instance, the ``step by step'' thinking guided by the CoT 
approach tends to make LLMs involve many 
correlation-based responses. The cluster approach of 
Auto-CoT also fails 
when the samples are too similar. Conversely, 
our iterative prompt instruction enable LLMs 
to uncover causal inconsistencies in its previous answers, 
thereby allowing for self-correction. This self-supervised 
idea appears to impose the LLMs with a capability of ``causal reasoning''.

Furthermore, to illustrate the specific progress of the LLM in each iteration, we select 4 cases to demonstrate their F1 dynamics during each iteration. Additionally, in order to explore the upper bound of F1, we conduct four types of supervision horizontally: unsupervised (n/a), meaning no specific causal relationships that need improvement are included in each feedback instruction; GPT2 supervision (GPT2), where GPT-2 is used to calculate $\hat{A}^{r}_{do_{g}}$ in step 2; GPT4 supervision (GPT4), which is the default setting of the algorithm; and label supervision (Label), where step 2 is skipped and the feedback is directly based on the conflict between the label and $\hat{A}^{s}_{do_{g}}$.  
\begin{table}
  \centering
  \resizebox{0.5\textwidth}{!}{
  \begin{tabular}{c|cc}
    \hline
    \bf Model&\bf Causalogue&\bf RECCON\\
    \hline
    Zero-Shot&0.61&0.52\\
    Zero-Shot-CoT&0.58&0.51\\
    Auto-CoT&0.62&0.51\\
    Ours&0.78&0.69\\
    \hline
  \end{tabular}}
  \caption{The F1 score of causal relationship recognition of different instructions.}
  \label{tabllm}
\end{table} 
\begin{figure*}
  \centering
  \subfigure[n/a]{
    \includegraphics[width=0.22\textwidth]{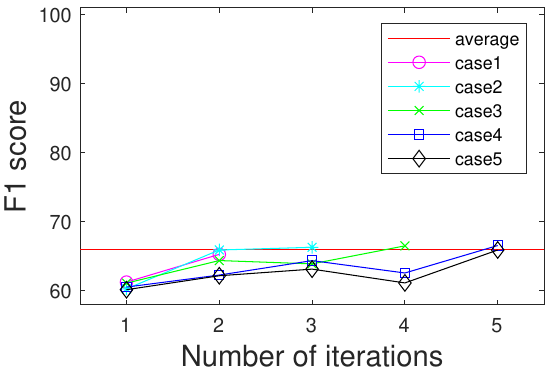}}
  \subfigure[GPT2]{
    \includegraphics[width=0.22\textwidth]{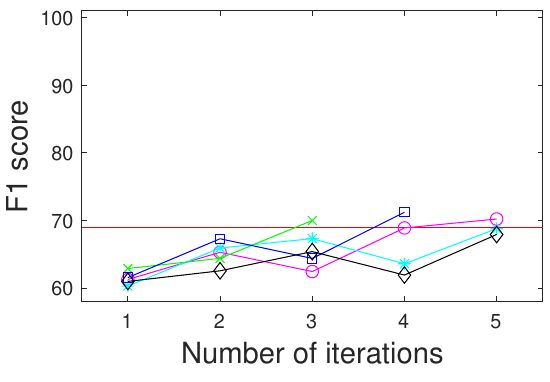}}
   \subfigure[GPT4]{
    \includegraphics[width=0.22\textwidth]{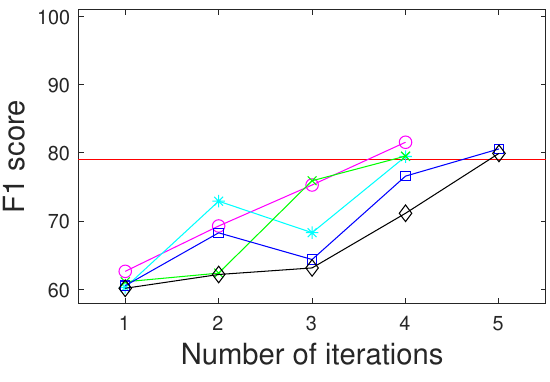}}
    \subfigure[Label]{
    \includegraphics[width=0.22\textwidth]{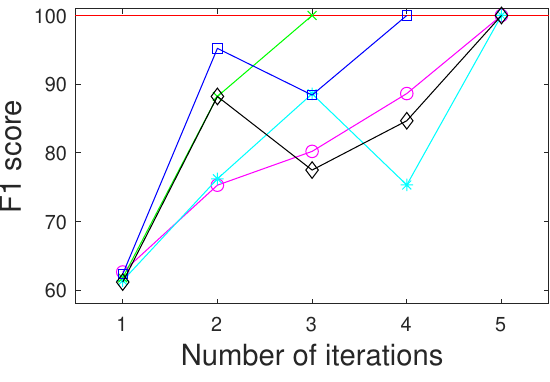}}
  \caption{F1 scores in each iteration of 5 cases with 4 different supervisions.}
  \label{figsupervision}
\end{figure*} 

Figure~\ref{figsupervision} confirms that the F1 value generally increases with the number of iterations, and the conditions for ending the iteration (no conflicts that can be fed back) are met in the 4th or 5th iterations. Moreover, different types of supervision can indeed affect the upper limit of the iteration, especially when conflicts between the label and the predicted results are directly fed back. The F1 value can ultimately approach 100. This suggests that feedback instructions may transform complex reasoning tasks such as ``causal relationship inference'' into simpler tasks that the LLM can handle (basic judgment and correction capabilities). Additionally, considering the results of GPT2 and GPT4, the inherent capabilities of the LLM are also an important factor in determining the upper limit of F1.

\section{Discussion}
\label{secdiscussion}

In fact, the inconsistency between causal representation and causal structure in Indefinite data fundamentally stems from the inability of high-dimensional representations to perfectly map into low-dimensional graph structure spaces. Starting with the simplest example, for high-dimensional representations of two variables $\hat{x}_{1}$ and $\hat{x}_{2}$, if they want to be consistent with the structure $x_{1}\rightarrow x_{2}$, this can easily be supervised and trained well (it just needs to supervise $f_{c}(\hat{x}_{1}, \hat{x}_{2})=1$ and $f_{c}(\hat{x}_{2}, \hat{x}_{1})=0$). However, for the high-dimensional representations of three variables $\hat{x}_{1}$, $\hat{x}_{2}$, and $\hat{x}_{3}$, if they want to be consistent with the structure $x_{1}\rightarrow x_{2}\rightarrow x_{3}$, it is much more difficult than with two variables. This is because it not only needs to ensure that $x_{1}\rightarrow x_{2}$ and $x_{2}\rightarrow x_{3}$ are supervised, but also that $x_{1}\rightarrow x_{3}$ is supervised. The challenge lies in the inability to directly obtain $f_{c}(\hat{x}_{1}, \hat{x}_{3})=1$ through $f_{c}(\hat{x}_{1}, \hat{x}_{2})=1$ and $f_{c}(\hat{x}_{2}, \hat{x}_{3})=1$, for example, when the structure $x_{1}\rightarrow x_{2}\rightarrow x_{3}$ and $x_{1}\rightarrow x_{4}\leftarrow x_{3}$ coexist. Therefore, as the number of variables increases, due to the need to consider the ``transitivity'' of the relationships between the variables in the structure, the ``supervision'' of high-dimensional representations explosively increases, making the correspondence between causal representation and structure increasingly difficult. Although SSL based on intervention can provide more constraints based on different views, it still cannot fundamentally solve the problem.

Furthermore, this kind of ``transitivity'' conflict between representation and structure is not only limited to the causal domain. For example, in the auto-circuit domain, there is always a conflict between the truth table of a single circuit representation and the truth table of multiple circuit representations. In the knowledge graph domain, there is a conflict between the representation relationship of a single hop and multiple hops. These works all suggest the need for a better mathematical model to transform the information of ``graph structure" into more sufficient constraint conditions.

\section{Conclusion}
\label{secconlusion}
In this paper, we focus on a frontier and novel research problem: the problem of causal inconsistency in causal structure and causal representation when using deep learning models for causal discovery in Indefinite Data. As the first work to identify this issue, this paper analyzes from both the model framework and empirical results perspectives how causal inconsistency arises in Indefinite Data and why it does not exist in other types of data. The crux of the problem lies in the backpropagation of causal structure and representation is conducted independently, making it challenging to maintain consistency in complex causal relationships. To mitigate this issue, we propose an intervention-based SSL framework, which provides ample supervision about causal relationships for both causal structure and representation through multiple intervention views. Extensive experiments demonstrate that the approach of enhancing consistency through intervention views can improve the performance of learning in causal structure and representation. Furthermore, we apply this framework to three practical downstream tasks and large language model scenarios, with the empirical results validating its practicality. Therefore, we believe that the intervention-based SSL framework is indeed an effective means to maintain causal consistency in complex causal data (such as Indefinite Data) during causal discovery.

\bibliography{iclr2025_conference}
\bibliographystyle{abbrvnat}

\appendix
\section{Follow-up Experiments for Consistency Test}
\label{secfollowupexperimentsonconsistency}
To test whether the addition of the loss function $distance(\hat{A}^{s}, \hat{A}^{r})$ could significantly enhance consistency, we introduced two indefinite datasets: Causalogue and Causaction~\citep{chen2024causal}. Moreover, for those methods applicable to indefinite data as listed in Table~\ref{tabinconsistency} (ACD, AVICI, DAG-GNN, CAE, biCD), we incorporated the new loss function $distance(\hat{A}^{s}, \hat{A}^{r})$ and retrained them. The training strategies and parameter settings were consistent with those described in their respective papers. For the ``$distance$", we adopted three types of distance metrics: cosine similarity (CS), Mean Squared Error (MSE), and Kullback-Leibler (KL) divergence. Table~\ref{tabdistancemearsure} displays the performance of these baseline models on the two datasets after the addition of the corresponding distance metrics. 

\begin{table*}
  \centering
  \resizebox{0.5\textwidth}{!}{
  \begin{tabular}{c|ccc|ccc}
    \hline
    \bf Methods&\multicolumn{3}{c}{\textbf{Causalogue}} \vline&\multicolumn{3}{c}{\textbf{Causaction}}\\
    &\textbf{stru}&\textbf{rep}&\textbf{inco}&\textbf{stru}&\textbf{rep}&\textbf{inco}\\
    \hline
    ACD&0.28&0.29&0.38&0.21&0.32&0.29\\
    +CS&0.26&0.25&0.37&0.22&0.31&0.29\\
    +MSE&0.28&0.28&0.37&0.21&0.28&0.28\\
    +KL&0.24&0.27&0.35&0.20&0.35&0.31\\
    \hline 
    AVICI&0.36&0.29&0.35&0.34&0.33&0.35\\
    +CS&0.36&0.28&0.36&0.29&0.32&0.34\\
    +MSE&0.33&0.31&0.34&0.31&0.38&0.38\\
    +KL&0.28&0.37&0.32&0.30&0.35&0.33\\
    \hline
    DAG-GNN&0.29&0.35&0.37&0.28&0.37&0.29\\
    +CS&0.34&0.29&0.34&0.29&0.35&0.36\\
    +MSE&0.34&0.29&0.35&0.37&0.31&0.28\\
    +KL&0.29&0.34&0.33&0.31&0.36&0.35\\
    \hline
    CAE&0.35&0.41&0.42&0.35&0.41&0.34\\
    +CS&0.35&0.34&0.36&0.37&0.39&0.39\\
    +MSE&0.36&0.34&0.35&0.33&0.41&0.38\\
    +KL&0.35&0.39&0.38&0.31&0.38&0.38\\
    \hline
    biCD&0.26&0.38&0.29&0.35&0.31&0.33\\
    +CS&0.28&0.34&0.33&0.29&0.31&0.32\\
    +MSE&0.29&0.31&0.33&0.31&0.38&0.34\\
    +KL&0.31&0.29&0.33&0.31&0.34&0.32\\
    \hline
  \end{tabular}}
   \caption{Comparison before and after adding distance metric. ``stru'', ``rep'', and ``inco'' are equal to Table~\ref{tabinconsistency}.}
  \label{tabdistancemearsure}
\end{table*}
Although different distance metrics have varying effects on different methods and datasets, overall, there is no universally significant improvement. In many scenarios, the performance even deteriorates. This suggests that simple distance metrics do not provide sufficiently strong constraints. Instead, they often exacerbate the difficulty of model convergence, outweighing the benefits of promoting consistency.

\section{Proof about equal intervention views leading to equal causal model} 
\label{suppproof}

To begin with, we would like to introduce causal factorization: 

The joint distribution of samples within the same structure 
can be represented by any factorization: 

\begin{equation}
   P(x_1,x_2,\dots,x_s)=\prod ^{S}_{s=1}P(x_{s}|X_{others})
   \label{eqafac}
\end{equation}
and it can always be consistent with the probability distribution 
of a certain graph. For instance, in statistical models, 
for any variable $x_j \in X_{others}$, there exists an 
undirected edge between $x_j$ and $x_i$, representing the 
correlation between $x_j$ and $x_i$. Despite the Markov property, 
the direction can't be directly identified by the 
conditional probability for an undirected edge. However, 
the causal model can identify the causal direction 
between two related variables $x_j$ and $x_i$ by intervention 
(for example, $x_j \rightarrow x_i$, $x_i \rightarrow x_j$, 
$x_j \rightarrow L \rightarrow x_i$, 
$x_j \leftarrow L \rightarrow x_i$ and so on). 

Exact transformation~\citep{rubenstein2017causal} 
 or causal abstraction~\citep{beckers2019abstracting} is a method  
 of judging causal consistency based on interventions and 
 SCMs. The $\tau $ transformation  becomes vital for making 
 two causal models equivalent. 

 \begin{definition}[$\tau$-transformation]
   Let $I_{L}$ to be a set of interventions on micro model 
   $SCM_{M}=\langle X_{M},\mathcal{F}_{M},U_{M},\mathbb{P}_{M}\rangle $. 
   Similarly, let $I_{N}$ be interventions on macro model 
   $SCM_{N}=\langle X_{N},\mathcal{F}_{N},U_{N},\mathbb{P}_{N}\rangle $. 
   Let $\tau$ be a partial transformation function 
   $\tau : \mathbb{P}_{M}(X_{M}) \rightarrow \mathbb{P}_{N}(X_{N})$. 
   Let $\omega : I_{M} \rightarrow I_{N}$ be 
   \begin{equation}
   \tau (\mathbb{P}_{M}(X_{M})) \rightarrow \mathbb{P}_{N}(X_{N})=\omega(I_{M})\rightarrow I_{N}
   \end{equation} 
\label{defta}
\end{definition} 

\begin{definition}[Causal Model]
   Let causal model 
   $M_{X}=\langle S_{X}, I_{X}, F_{X}^{I_{X}}\rangle $, 
   where $S_{X}$ represents an SCM for the model with the variable 
   set $X=(x_{i}:i\in \zeta_{x})$, $\zeta_{x}$ is the index of 
   causal partial order over $X$, 
   $I_{X}:=({do(i):i\in \zeta_{x}})$ 
   represents a set of all reasonable bi-variable perfect 
   interventions satisfying partial order, 
   $F_{X}^{I_{X}}:=({f^{do(i)}_{X}:i\in \zeta_{x}})$
   represents the causal strength of set $X$ under corresponding 
   interventions. 
   \label{defcm}
   \end{definition} 

Existing work~\citep{hu2022neuron} proposed that two causal models 
satisfying causal abstraction are consistent. Accordingly, 
we employ causal abstraction as a mediator.  

First, we would like to prove that equivalent distribution sets 
$\rightarrow$ equivalent strength sets. 

According to the causal factorization, $P(x_{s}|X_{others})$ 
satisfies $x_{s}=f_{s}(Pa(x_{s},u_{s}))$. When we assume two 
causal model within two variables: 
$\mathcal{U}:x_{i}\rightarrow x_{j}$ and 
$\mathcal{V}:y_{i}\rightarrow y_{j}$, we would like to adopt 
SCM to represent the equivalent distribution 
$P_{x_{j}}(do(x_i))=P_{y_j}(\omega (do(x_i)))$: 
\begin{equation}
   f_{i,j}(u_{x_i})+u_{x_j}=g_{i,j}(u_{y_i})+u_{y_j}
\label{eqap1}
\end{equation} 

Because of $P_{x_{j}}(do(x_j))=P_{y_j}(\omega (do(x_j)))$  and 
$P_{x_{i}}(do(x_i))=P_{y_i}(\omega (do(x_i)))$, Equation~\ref{eqap1} 
can be written as: 

\begin{equation}
   f_{i,j}=g_{i,j}
\label{eqap2}
\end{equation} 

According to the causal partial order, 
$P_{x_{j}}(do(x_j)) \preccurlyeq P_{x_{j}}(do(x_j,x_k)), (x_j \preccurlyeq x_k)$, 
hence:

\begin{equation}
   f_{j,k}=g_{j,k}
\label{eqap3}
\end{equation} 
When we convert any causal factorization into a chain of 
ancestral relationships through additive noise formulas, 
it is possible to find a corresponding $g_{i,j}$ that equals 
$f_{i,j}$ for any step in the causal chain. 
Finally, we can infer that if the distribution sets 
$P_X = P_Y$, then the strength sets $F_X = F_Y$. Conversely, 
it can also be proven that if the strength sets $F_X = F_Y$, 
then the distribution sets $P_X = P_Y$. 

 Finally, we 
would like to introduce the causal consistency condition: 

\begin{theorem}[Causal Consistency Condition (CCC)]
   Let $\mathcal{U}_{X}=(S_{X},I_{\ast},F^{I_{\ast}}_{X})$ and 
   $\mathcal{V}_{Y}=(S_{Y},I_{\ast},F^{I_{\ast}}_{Y})$ be two 
   causal models. The intervention set $I_{\ast}$ denotes that 
   there is an identity mapping between $X$ and $Y$. If any term 
   $f_{y_1,y_2}^{do(i)}$ in $F^{I_{\ast}}_{Y}$ satisfies: 
   \begin{equation}
      f_{y_1,y_2}^{do(i)}=f_{x_1,x_2}^{do(i)}
   \end{equation}
   the $\mathcal{U}_{X}$ is consistent with $\mathcal{V}_{Y}$ 
   \label{thmccc}
\end{theorem}

\section{Details about Datasets, Metrics, Baselines, and Implementation}
\label{suppddbi}

\subsection{Datasets}
Their data splits and specific $N$-folds validation setups for 
SSM are exhibited in Table~\ref{tabsd}. 
Among them, \textit{Causalogue}, RECCON, DD, MELD, EmoryNLP and IEM are text datasets, 
and \textit{Causaction}, GTEA, 50salads, and Breakfast are video datasets. 
As for LLM, we only randomly select 400 samples from 
\textit{Causalogue} and RECCON datasets, respectively. 
Their overviews and prevalent metrics are detailed below. 
\begin{table}
  \centering
  \resizebox{0.5\textwidth}{!}{
  \begin{tabular}{ccccc}
    \hline
    \bf Dataset&\bf Train&\bf Valid&\bf Test&\bf Folds\\
    \hline
    Causalogue&1338&100&200&10\\
    Causaction&818&100&200&5\\
    RECCON&833&47&225&10\\
    DD&11118&1000&1000&5\\
    MELD&1038&114&280&5\\
    EmoryNLP&713&99&85&5\\
    IEM&100&20&31&5\\
    GTEA& 19& 2& 7& 10\\
    50salads&36 &4 &10 &10 \\
    Breakfast&1314 &146 &252 & 10\\
    \hline
  \end{tabular}}
  \caption{Statistics on Datasets}
  \label{tabsd}
\end{table}

\subsubsection{Causal Discovery in Indefinite Data (CDID) Task}
\textbf{Causalogue}~\citep{chen2024causal}: is the first dialogue 
dataset that includes comprehensive causal relationship labels 
for Indefinite Data. It employs GPT-4 generation as a
substitute for data collection from the real world or manual
simulation. The dataset incorporates 10 types of causal structures (M
= 10), each with 44-276 dialogues, with each one comprised of 
4 utterances define as 4 causal variables.  

\textbf{Causaction}~\citep{chen2024causal} is another Indefinite Dataset 
obtained after re-annotating the Breakfast Dataset~\citep{kuehne2014language}.
It contains a total of 1,118 videos, documenting 10 different
breakfast preparation processes (such as coffee, salad, sandwich, etc.). 
Each video consists of 4-9 actions, with a clear
frame boundary. The causal relationship between any two actions 
are annotated by human. 

\subsubsection{Emotion-cause Pair Extraction (ECPE) Task}
\textbf{RECCON}~\citep{poria2021recognizing}: 
The first dataset for emotion cause recognition of conversation 
including RECCON-DD and RECCON-IE (emulating an out-of-distribution 
generalization test). RECCON-DD includes 5380 labeled ECPs and 5 cause 
spans (\textit{no-context}, \textit{inter-personal}, \textit{self-contagion}, 
\textit{hybrid}, and \textit{latent}). 

\subsubsection{Emotion Recognition in Conversation (ERC) Task}
\textbf{DD}~\citep{li-etal-2017-dailydialog}:
A Human-written dialogs dataset with 7 emotion labels (\textit{neutral}, 
\textit{happiness}, \textit{surprise}, \textit{sadness}, 
\textit{anger}, \textit{disgust}, and \textit{fear}). We follow 
~\citep{chen2023affective} to regard utterance turns as the speaker turns. 

\textbf{MELD}~\citep{poria-etal-2019-meld}: 
A multi-model ERC dataset with 7 emotion labels as the same as DD. 

\textbf{EmoryNLP}~\citep{zahiri:18a}:
A TV show scripts dataset with 7 emotion labels (\textit{neutral}, 
\textit{sad}, \textit{mad}, \textit{scared}, \textit{powerful}, 
\textit{peaceful}, \textit{joyful}). 

\textbf{IEM}~\citep{busso2008iemocap}: 
A multi-model ERC dataset with 9 emotion labels (\textit{neutral}, 
\textit{happy}, \textit{sad}, \textit{angry}, \textit{frustrated}, 
\textit{excited}, \textit{surprised}, \textit{disappointed}, and \textit{fear}). 
However, models in ERC field are often evaluated on samples with 
the first six emotions due to the too few samples of the latter three emotions. 
20 dialogues for validation set is following~\citep{chen2023affective}. 

\subsubsection{Temporal Action Segmentation (TAS) Task}
\textbf{GTEA}~\citep{fathi2011learning} 
Georgia Tech Egocentric Activities is comprised of 28 videos 
captured from a first-person perspective. It documents 
7 different daily activities performed by 4 test actors, 
therefore, the dataset is partitioned into four 4 
based on the actors. Each video contains approximately 
20 fine-grained instances, with each video divided by 
action segments as labels. 

\textbf{50salads}~\citep{stein2013combining} 
A cooking dataset includes 50 videos highlighting the 
complete process of salad preparation undertaken by 25 people, 
with each video housing between 9,000 to 18,000 RGB frames 
and containing 17 action class labels. Each video, 
named after the complete process of salad making by an individual, 
is segregated into 5 groups. 

\textbf{Breakfast} ~\citep{kuehne2014language}
A cooking action dataset consists of 10 cooking activities 
performed by 52 different actors at various kitchen locations. 
It encompasses 1,989 videos and offers over 77 hours of content. 
Each video is characterized by a sub-cooking activity 
accomplished by an actor; the complete preparation process 
comprises 20-30 such action segments. As the largest 
among the mentioned datasets, it is divided into 4 groups. 

\subsection{Evaluation Metrics}\label{ssem}
\subsubsection{Causal Discovery in Indefinite Data (CDID) Task}
The CDID task was evaluated on the \textit{Causalogue} 
and \textit{Causaction} dataset. In our experiments, 
we endeavored to assess three outcomes: 
the accuracy of causal graphs, the accuracy of causal 
representations, and the consistency between causal graphs 
and representations. Consequently, we employed AUROC and 
Hamming Distance (HD) to measure causal graphs, 
AUROC and F1 scores for causal representation evaluation, 
and MSE and $1-$AUROC for measuring the distance of inconsistencies. 
These metrics are common and well-accepted. 
Simultaneously, for each outcome, we ensured two different 
metrics to comprehensively evaluate the performance. 

\subsubsection{Emotion-cause Pair Extraction (ECPE) Task}
We continue to employ the F1 score as the evaluation metric, 
as initially proposed in~\citep{poria2021recognizing}. 
This metric is broadly accepted and utilized in current 
research works~\citep{li2023experiencer,li2023class}. 

\subsubsection{Emotion Recognition in Conversation (ERC) Task} 
Similarly to ECPE task, We continue to employ the F1 score 
as the evaluation metric, as initially proposed 
in~\citep{shen-etal-2021-directed}. 
This metric is broadly accepted and utilized in current 
research works~\citep{chen2023affective,zhang2023dualgats,
shi2023multiemo}. 
\subsubsection{Temporal Action Segmentation (TAS) Task}\label{ssstas}
Commonly used metrics include frame-level accuracy (Acc), 
segmental edit distance (Edit), and segmental F1 scores 
with different overlapping threshold k (F1$@$k) ($k =\{10, 25, 50\}$). 
Moreover, to evaluate the causal consistency  
of the segmentation results, we proposed an additional causal 
edit distance (C-Dis) to measure the dissimilarity between the 
adjacency matrix and the ground truth. For the final segmentation 
results, we constructed causal adjacency matrices 
$\hat{C} \in \mathbb{R}^{T*T}$ and ground truth matrices 
$C \in R^{T*T}$, based on the constraints in consistent 
mapping condition and calculated the dissimilarity between them. 
\begin{equation}
C-Dis :=num(\hat{C}_{i,j}  \neq  C_{i,j}) for~ i,j=1,2,…,T 
\end{equation}
A lower causal edit distance indicates that the causal 
relationship at the frame-level has less dissimilarity 
with the ground truth, 
demonstrating stronger learning ability with 
causal representation in the model, 
and hence a higher level of causal consistency 
in the segmentation results. 
\subsection{Baselines}

\subsubsection{Baselines on CDID Task}

\textbf{Supervised Specialized Model (SSM)}

\textbf{ACD}: leverages shared dynamics to learn to infer causal 
relationships from multi-structure time-series data via a 
single, amortized model. 

\textbf{DAG-GNN}: leverages SCM to construct a gnn-based variational 
model adopting independent noise $E$ as latent variable. 

\textbf{ACCD}: discover causal relationships in multi-value data 
via designing a common structure and generating a substitute for 
independent noise.

\textbf{biCD}: proposes a dynamic variational inference model leveraging 
the causal strength instead of independent noise as the latent variable 
to construct ELBO for Indefinite Data. 

\textbf{DisC}: designs a new method for intervention in deep models, 
combining causal patterns with different shortcuts to achieve 
the goal of intervention in causal nodes.  

\textbf{DIR}: distinguishes between positive and negative samples 
after intervention by designing a dynamic loss function, 
Similar to the DisC 
thereby effectively intervening in the causal pattern. 

Since DisC and DIR do not have complete causal discovery models, 
we incorporate their intervention modules into DAG-GNN 
(namely, DAG-DisC and DAG-DIR) 
to demonstrate their intervention strategies for Indefinite Data. 

\textbf{Large Language Model (LLM)}

\textbf{Zero-shot and Zero-shot-CoT}: proposes a new prompt paradigm 
like ``Let's think step by step'' which is task-agnostic and does 
not need input-output demonstrations. 

\textbf{Auto-CoT}: proposes an auto prompt method which could 
cluster the samples first and then select an example for prompt text.

\subsubsection{Baselines on ECPE Task}

\textbf{EDKA-GM}: introduces an experiencer identification task 
and present a document-level heterogeneous graph network 
for capturing global experiencer information to 
enrich experiencer-based cross-clause association. 

\textbf{seF}: includes two main components: 
core clause selector and emotion-cause pairs extractor to 
jointly extract emotion-cause pairs. 

\subsubsection{Baselines on ERC Task}
\textbf{DAG-ERC}: proposes a gnn$\&$rnn-based model to learn the 
relationship of different speakers and sequential information. 

\textbf{DualGAT}: introduces Dual Graph Attention networks to 
concurrently consider the complementary aspects of discourse 
structure and speaker-aware context. 

\textbf{MultiEMO}: proposes a novel attention-based 
correlation-aware multimodal fusion framework 
effectively integrating multimodal cues by capturing 
cross-modal mapping relationships across textual, 
audio and visual modalities. 

\subsubsection{Baselines on TAS Task}

\textbf{MS-TCN}: This is the first method to introduce a 
multi-stage action segmentation framework based on 
Temporal Convolutional Networks (TCN). Each stage inputs 
the initial prediction output from the preceding one 
for further modification and adjustment. 

\textbf{MS-TCN++}: On the foundation of MS-TCN, 
this method introduces a dual dilated layer, 
implementing parameter sharing and optimizing segmentation 
performance. 

\textbf{ASRF}: This method proposes an improved technique based 
on MS-TCN, composed of a long-term feature extractor 
and two branches: the Action Segmentation Branch (ASB) 
and the Boundary Regression Branch (BRB). 

\textbf{CETNet}: Leveraging Transformer, this method connects 
every layer of convolutional feature mapping in the encoder 
with a group of features generated through self-attention 
in the decoder. 

\textbf{C2F}: Utilizing TCN, this method puts forward a 
novel temporal encoder-decoder to tackle the sequence fragment 
issue. Its decoder conforms to a coarse-to-fine structure 
with multi-timescale implicit integration. 

\subsection{Implementation Details}\label{secid}

\subsubsection{The Model on CDID Task}
In our Experiments, we utilized RoBERTa-base (768) (for \textit{Causalogue}) 
and I3D (for \textit{Causaction}) 
as our pre-trained model for generating input representation 
in the SSMs. Throughout the training process, for 2 datasets separately,  
a learning rate of 1e-4/1e-5 was set, 
with the batch size and epochs set to 4/2 and 50, respectively. 
The dimension of the hidden layers within the network was also set 
to 768/1024. The entire training procedure was conducted on a 
NVIDIA GEFORCE RTX 3090 graphics processing unit. In the ablation study, we replace the encoder and decoder with learnable distributions. Specifically, we utilize the encoder and decoder structures from~\citep{chen2023affective} as substitutes.

\subsubsection{The Model on ECPE Task and ERC Task} 
In the word embedding, we adopt the affect-based pre-trained 
features\footnote{\url{https://drive.google.com/file/d/1R5K_2PlZ3p3RFQ1Ycgmo3TgxvYBzptQG/view?usp=sharing}} 
proposed by~\citep{shen-etal-2021-directed} for all baselines and models. 

In the hyper-parameters, we follow the setting of~\citep{chen2023affective} 
in the ERC task. Moreover, in the ECPE, the learning rate is set to 3e-5, 
batch size is set to 32, and epoch is set to 60. Further in our approach, 
hidden size of GNN is set to 300, and dropout rate is 0.3. 
All experiments were conducted on a NVIDIA GEFORCE RTX 3090 
for both training and testing. 

\subsubsection{The Model on TAS Task}
We employed the features extracted from I3D~\citep{carreira2017quo}
as the input for our model. To avoid random bias, 
we applied our augmented approach across different backbones 
while retaining the seed setup from their original studies, 
ensuring that the specific training epochs are consistent 
with the backbones. All experiments were conducted on a 
NVIDIA GEFORCE RTX 3090 for both training and testing. 
We set the learning rate to 0.0005, established a weight decay 
of 0.001, and utilized Adam as the optimizer. 
To enhance the training efficiency and avert degenerate 
matrices during whitening, we set the batch size of the 
frame images for a video segment to 512. We followed 
the recommendation in~\citep{ermolov2021whitening} 
to further subdivide the batches during the whitening process, 
setting the sub-batch size to 128. 

\section{Other Experiments of Causal Consistency} 

\subsection{Variance of Learning Results}
\label{expvlr}
\begin{figure}
  \centering
  \includegraphics[width=0.6\textwidth]{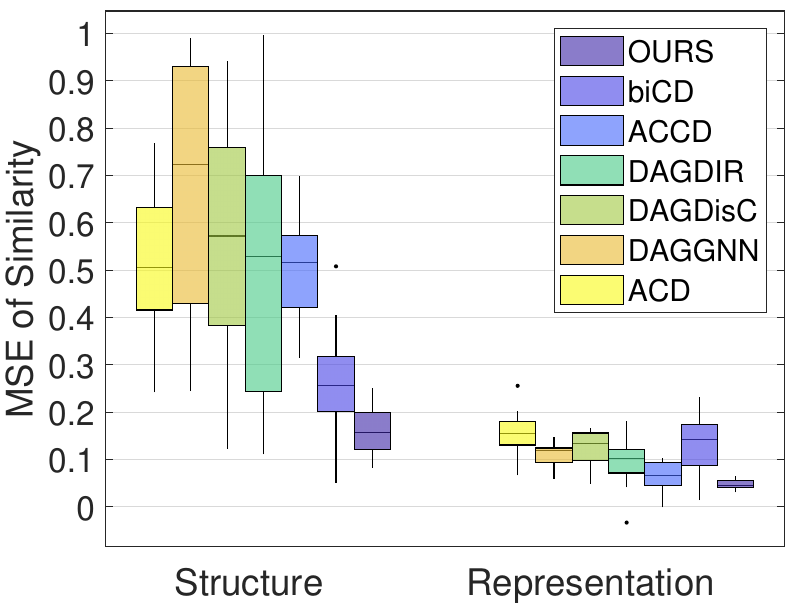}
  \caption{The boxplot showing the MSE of predicted $\hat{A}$ to ground truth. The left clustering is the results 
  from structure ($\hat{A}^{s}$) to the ground truth, and right clustering is results 
  from representation ($\hat{A}^{r}$) to the ground truth.}
  \label{figvariance}
\end{figure}

Figure~\ref{figvariance} 
shows the box plots of the errors in structure and representation. 
Ours evidently outperforms other methods in terms 
of structure, even those specifically designed to handle 
complex variables (ACCD, biCD). In addition, the variance 
of intervention-based methods (DAG-DisC, DAG-DIR) 
is extremely large, which aligns with our previous conclusion 
that intervening by negative examples leads to the additional 
variance from the batch size. 
On the representation side, almost all methods performed well. 
Nearly all methods can reduce the error to a certain extent. 
Nevertheless, the remaining stubborn error is due to 
"pseudo-correlation" caused by the inability to fully satisfy 
causal relationships due to causal inconsistency. 
The significantly smaller variance of our method demonstrates 
that intervention views can further help representation more completely determine 
causal relationship. 

\subsection{Scalability}
\label{expscalability}
\begin{table}
  \centering
  \resizebox{\textwidth}{!}{
  \begin{tabular}{cccccccccccccccc}
    \hline
    \bf Methods&\multicolumn{3}{c}{\textbf{20$\%$ Trainset}}&\multicolumn{3}{c}{\textbf{40$\%$ Trainset}}&\multicolumn{3}{c}{\textbf{60$\%$ Trainset}}&\multicolumn{3}{c}{\textbf{80$\%$ Trainset}}&\multicolumn{3}{c}{\textbf{100$\%$ Trainset}}\\
    &auroc$_{S}$&auroc$_{R}$&auroc$_{C}$&auroc$_{S}$&auroc$_{R}$&auroc$_{C}$&auroc$_{S}$&auroc$_{R}$&auroc$_{C}$&auroc$_{S}$&auroc$_{R}$&auroc$_{C}$&auroc$_{S}$&auroc$_{R}$&auroc$_{C}$\\
    \hline
    ACD         &0.15&0.22&0.02              &0.31&0.30&0.19              &0.41&0.38&0.31               &0.49&0.45&0.44               &0.55&0.55&0.51\\
    DAG-GNN     &0.08&0.16&0.06              &0.25&0.27&0.25              &0.31&0.34&0.33               &0.34&0.38&0.45               &0.41&0.50&0.50\\
    DAG-DisC    &0.07&0.24&0.08              &0.41&0.32&0.24              &0.50&0.44&0.31               &0.54&0.51&0.46               &0.58&0.58&0.52\\
    DAG-DIR     &0.10&0.26&0.14              &0.36&0.37&0.29              &0.46&0.48&0.28               &0.53&0.54&0.46               &0.57&0.59&0.51\\
    ACCD        &0.13&0.29&0.12              &0.29&0.41&0.27              &0.35&0.52&0.39               &0.41&0.58&0.55               &0.46&0.63&0.60\\
    biCD        &0.16&0.25&0.18              &0.33&0.39&0.26              &0.54&0.51&0.61               &0.57&0.58&0.81               &0.66&0.65&0.89\\
    \hline
    Ours$_{SSM}$&\bf 0.21&\bf 0.34&\bf 0.18  &\bf 0.42&\bf 0.41&\bf 0.35  &\bf 0.55&\bf 0.52&\bf 0.69   &\bf 0.62&\bf 0.61&\bf 0.89   &\bf 0.69&\bf 0.67&\bf 0.95\\
    \hline
  \end{tabular}}
  \caption{Results of Scalability on \textit{Causalogue} Dataset. 
  ``auroc$_{S}$'', ``auroc$_{R}$'', and ``auroc$_{C}$'' represent the AUROC of 
  Causal Structure, Causal Representation and Causal Consistency, respectively. }
  \label{tabscalability}
\end{table}

We evaluate scalability by scaling the training set. 
Table~\ref{tabscalability} shows that our method performs best 
under any scale of datasets, especially in terms of structure. 
This is because, when the sample size is insufficient, 
intervention methods can extract more causal information 
contained in the samples.




\end{document}